\documentclass[]{style}

\usepackage{graphicx}
\usepackage{float}
\usepackage{amsmath}
\usepackage{amsfonts}
\usepackage{amssymb}
\usepackage{multirow}
\usepackage{colortbl}
\usepackage{threeparttable}
\usepackage{booktabs}
\usepackage{adjustbox}
\usepackage[misc]{ifsym}
\usepackage{xcolor}
\usepackage{makecell}
\usepackage{pifont}
\usepackage{xspace}

\renewcommand{\emph}[1]{\textit{#1}}

\providecommand{\IEEEPARstart}[2]{#1#2}
\newcommand{\cmark}{\ding{51}}
\newcommand{\xmark}{\ding{55}}

\makeatletter
\DeclareRobustCommand\onedot{\futurelet\@let@token\@onedot}
\def\@onedot{\ifx\@let@token.\else.\null\fi\xspace}

\makeatother

\newcommand{\papertitle}{Open-o3-Video-v2}

\title{
  \papertitle{}: Seek, Ground, and Reason with Verifiable Spatio-Temporal Evidence
}

\author[]{Open-o3-Video Team: PKU, FDU, NTU, UC Merced}

\abstract{Reliable video reasoning requires models to produce correct answers and expose supporting visual evidence. Open-o3-Video enables verifiable reasoning by embedding timestamps and object-level bounding boxes into reasoning traces, but is restricted to frames from an initial sparse sampler.
We introduce Open-o3-Video-v2, which formulates video reasoning as a unified agentic evidence-grounded process. The model identifies missing information, acquires additional observations, grounds the moments and objects supporting its reasoning, and writes this evidence into the trajectory before continuing or answering.
To support this behavior, we construct STGR-v2-CoT-135K, including 52K multi-turn agentic grounded reasoning trajectories, and STGR-v2-RL-13K for reinforcement learning. We further propose Clue-Seek GRPO (CS-GRPO), which tracks how tool calls change accumulated evidence, rewards clue coverage and step-wise gains, and penalizes excessive, repetitive, or uninformative retrieval.
On V-STaR, Open-o3-Video-v2-4B achieves 40.4\% mAM and 56.3\% mLGM, outperforming its Qwen3-VL-4B backbone by 8.5\% and 12.6\%, respectively. It also delivers consistent gains across general video understanding, long-video reasoning, and temporal grounding benchmarks.
These results show that active evidence acquisition and verifiable object-level grounding are complementary for reliable video reasoning.
The code and models are released at \url{https://msalab-pku.github.io/projects/Open-o3-Video-v2}.

}

\begin{document}
\maketitle

\section{Introduction}
\label{sec:intro}

\begin{figure*}[t]
  \centering
  \includegraphics[width=1.0\textwidth]{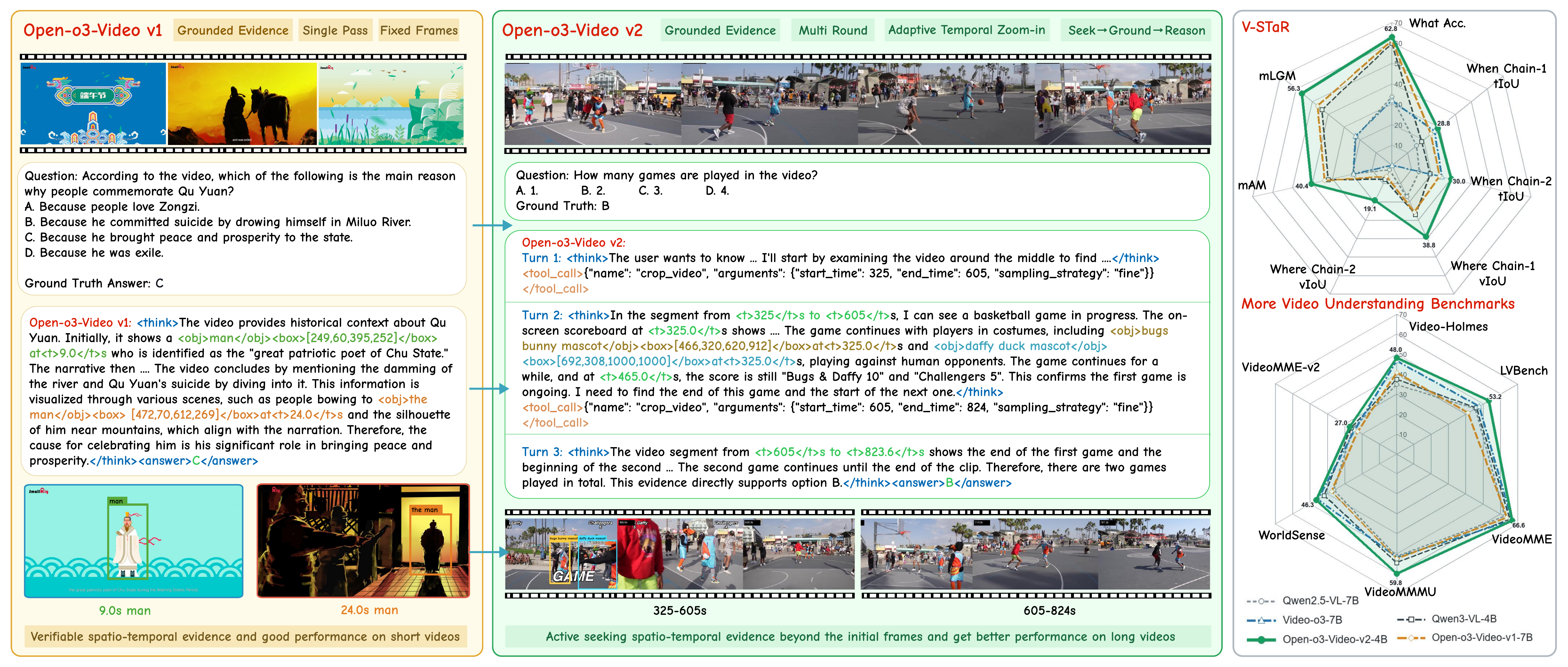}
  \vspace{-2mm}
  \captionof{figure}{\textbf{From single-pass spatio-temporal grounded reasoning to agentic evidence seeking and grounded reasoning.}
  \emph{Left:} Open-o3-Video-v1 reasons over fixed sampled frames and grounds visible evidence, yielding verifiable spatio-temporal support on short videos.
  \emph{Center:} Open-o3-Video-v2 performs adaptive temporal zoom-in through
  iterative \texttt{crop\_video} calls, following a
  \emph{seek--ground--reason} process to find evidence beyond the initial frames for long video understanding. \emph{Right:} Comparison of five models on V-STaR and six general video understanding and reasoning
  benchmarks.}
  \label{fig:teaser}
  \vspace{-1em}
\end{figure*}

\IEEEPARstart{V}{ideo} understanding has advanced rapidly with multimodal large language models (MLLMs)~\cite{qwen3.5,bai2025qwen3vl,zhang2025videollama3,an2026llava}. These models can summarize long videos, answer questions about dynamic events, and localize relevant segments. Post-training methods~\cite{feng2025video-r1,wang2025videorft} further strengthen temporally aware reasoning through chain-of-thought supervision and reinforcement learning. Yet reliable, fine-grained reasoning over long and cluttered scenes remains difficult. Rapid motion, frequent scene changes, and subjects repeatedly leaving and reentering the view make visual events difficult to track, while a decisive clue may occur briefly or occupy only a small region. These settings demand more than fluent language reasoning: a capable model must locate reliable visual evidence and compose it into a defensible inference.
Recent work on \textit{Thinking with Images}~\cite{openai-o3,wang2025traceable} offers a promising direction. By interleaving language with operations such as cropping and zooming, these methods let a model inspect details and feed new visual information back into reasoning. Extending this paradigm to video raises a central question: how can a model integrate visual observations into its reasoning so that both the intermediate process and final answer are reliable, accurate, and verifiable?

\begin{table}[t]
  \centering
  \caption{Capability comparison with closely related video reasoning methods. ``Adaptive temp. res.'' denotes adaptive temporal resampling, ``Temp. ev.'' temporal evidence explicitly cited in reasoning, ``Obj. ev.'' object-level evidence explicitly cited in reasoning, and ``Native agent'' multi-turn interaction with interleaved observations.}
  \label{tab:method_capability_comparison}
  \setlength{\tabcolsep}{2.7pt}
  \scriptsize
  \begin{tabular}{lcccc}
    \toprule
    \textbf{Method} & \makecell{\textbf{Adaptive temp. res.}} & \textbf{Temp. ev.} & \textbf{Obj. ev.} & \textbf{Native agent} \\
    \midrule
    Open-o3-Video~\cite{meng2025open} & \xmark & \cmark & \cmark & \xmark \\
    VideoZoomer~\cite{ding2025videozoomer} & \cmark & \xmark & \xmark & \cmark \\
    Video-o3~\cite{zeng2026video} & \cmark & \xmark & \xmark & \cmark \\
    VideoTemp-o3~\cite{liu2026videotemp} & \cmark & \xmark & \xmark & \cmark \\
    VideoSeeker~\cite{zhao2026videoseeker} & \cmark & \xmark & \xmark & \cmark \\
    \midrule
    \textbf{Open-o3-Video-v2} & \cmark & \cmark & \cmark & \cmark \\
    \bottomrule
  \end{tabular}
  \vspace{-2em}
\end{table}

Achieving grounded and active video reasoning is difficult for three reasons.
\textnormal{First}, suitable supervision is scarce and fragmented: video question-answering datasets typically omit visual evidence, grounding datasets provide only partial signals, and few resources jointly capture temporal clues, object localization, multi-turn tool interactions, and aligned reasoning traces.
\textnormal{Second}, spatio-temporal grounding must align with the chain of thought: plausible timestamps, boxes, and rationales may still be mutually inconsistent, while final-answer supervision or independently optimized grounding signals do not ensure that each reasoning claim is supported by the correct moment and object. Reliable optimization therefore requires structured, evidence-aware supervision linking intermediate reasoning to visual support.
\textnormal{Third}, a single observation is often insufficient: sparse sampling may miss brief actions or small objects, whereas dense processing is costly and introduces irrelevant context. Agentic frame selection and temporal cropping~\cite{zeng2026video,ding2025videozoomer,zhao2026videoseeker} offer an adaptive alternative, but retrieved clips are often weakly tied to fine-grained object evidence, and task-level rewards provide limited credit for whether each call contributes useful new information. Effective tool use should therefore account for how successive observations update the accumulated evidence for reasoning.

To address these challenges, we develop two generations of \textbf{Open-o3-Video} for reasoning with verifiable visual evidence. The original \textbf{Open-o3-Video} embeds timestamps and object boxes directly into the chain of thought, making fine-grained spatio-temporal reasoning traceable~\cite{meng2025open}. It introduces \textbf{STGR-CoT-30K} and \textbf{STGR-RL-36K} to jointly supervise temporal grounding, spatial grounding, general video QA, and spatio-temporal grounded reasoning. Each grounded trace links semantic reasoning claims to timestamped frames and localized objects, making intermediate visual evidence readable, parseable, and directly optimizable beyond final-answer correctness. Full construction details are provided in Appendix~\ref{app-sec:v1_details}.

Open-o3-Video addresses the coupled optimization of temporal and spatial grounding. Since spatial rewards are meaningful only after the relevant moment is identified, inaccurate early timestamps can suppress useful spatial feedback. \textbf{Adaptive temporal proximity} gradually tightens the temporal reward neighborhood during training, while \textbf{temporal gating} activates spatial rewards only for sufficiently accurate timestamps. Together, they alleviate early reward sparsity while preventing boxes at irrelevant moments from receiving credit. Open-o3-Video consequently improves mAM and mLGM over Qwen2.5-VL by 14.4\% and 24.2\% on V-STaR~\cite{VSTaR}, with consistent gains on multiple general video and temporal-grounding benchmarks. Detailed ablations are reported in Appendix~\ref{app-sec:v1_ablation}.
Nevertheless, Open-o3-Video remains restricted to its initially sampled frames and cannot recover decisive evidence omitted by that observation.

Going beyond this fixed visual context, we develop \textbf{Open-o3-Video-v2} by formulating video reasoning as a unified agentic evidence-grounded process.
As summarized in Table~\ref{tab:method_capability_comparison}, recent agentic video methods allow models to retrieve targeted temporal observations. However, acquiring a relevant interval mainly determines \emph{where to observe}. Reliable evidence-grounded reasoning further requires the model to determine which moments and objects within the acquired observation support a particular inference.
Open-o3-Video-v2 integrates these decisions into the same reasoning trajectory. The model first identifies unresolved information and decides where additional observation is needed. It then grounds the acquired observation to query-relevant moments and objects and records this evidence in the reasoning history. The newly acquired evidence becomes an explicit premise for subsequent inference, allowing the model to revise its interpretation, continue seeking, or answer from the accumulated evidence.
As illustrated in Fig.~\ref{fig:teaser} (middle), this seek--ground--reason process turns tool-acquired observations into explicit evidence that directly participates in the evolving reasoning process. The central contribution of v2 is therefore not temporal cropping or grounding in isolation, but their integration into an agentic reasoning formulation in which evidence is actively acquired, grounded, and reused.
Our main extensions to Open-o3-Video and contributions are threefold:

\textbf{First}, we establish an \textbf{agentic evidence-grounded reasoning formulation} that extends grounded reasoning from fixed observations to an iterative process of seeking, grounding, and integrating missing evidence. The model jointly determines where additional observation is required and which moments and objects in the acquired content provide support for subsequent reasoning.
\textbf{Second}, we construct \textbf{STGR-v2-CoT-135K}, extending grounded supervision from single responses to multi-turn agentic trajectories. Its core 52K examples connect sparse observation, targeted video cropping, returned frames, object-level grounding, and final answers in coherent reasoning traces. To encourage evidence-driven rather than ritualized tool use, we remove questions consistently solved from sparse observations, verify that each retained clue supports the answer, annotate timestamped object boxes with SAM3~\cite{carion2025sam3}, and filter inconsistent calls, boxes, and visual references. Together with grounding, dense-captioning, general video QA, and auxiliary tool-use data, the resulting corpus contains 135K SFT samples. We further construct \textbf{STGR-v2-RL-13K} with informative rollout samples for subsequent policy optimization.
\textbf{Third}, we propose \textbf{Clue-Seek GRPO (CS-GRPO)} to optimize evidence acquisition over multi-turn trajectories. CS-GRPO tracks how the accumulated temporal clue state changes across successive interactions, rewards final clue coverage and useful evidence gains, and regularizes excessive, repetitive, fragmented, or uninformative retrieval. Together with answer and grounded-thinking supervision, the final objective jointly optimizes evidence acquisition and the grounding of acquired observations.

Open-o3-Video-v2-4B achieves 40.4\% mAM and 56.3\% mLGM on V-STaR, improving the Qwen3-VL-4B baseline by 8.5\% and 12.6\%, respectively. It also improves over the baseline across general video understanding and temporal-grounding benchmarks, including gains of 10.7\% on Video-Holmes~\cite{cheng2025videoholmes} and 7.2\% on LVBench~\cite{LVBench}. The 8B model further reaches 56.5\% mLGM on V-STaR, 63.0\% accuracy on VideoMMMU, and 51.1\% mIoU on Charades-TimeLens~\cite{zhang2026timelens}.

\section{Related Work}
\label{sec:related_work}

\noindent\textbf{Video Spatial and Temporal Grounding.}
Video grounding identifies when an event occurs and where its visual referents appear. For temporal grounding, TimeChat~\cite{ren2023timechat} associates frames with timestamps and uses a sliding video query transformer, while LITA~\cite{huang2024lita} combines relative time tokens with SlowFast temporal modeling. Grounded-VideoLLM~\cite{wang2024grounded} preserves inter-frame dynamics through an additional temporal stream and discrete time tokens, and UniTime~\cite{li2025unitime} generates interleaved timestamps with coarse-to-fine localization. Recent methods directly optimize boundary quality with verifiable rewards: Time-R1 and TVG-R1~\cite{wang2025time,chen2025datasets} reward temporal overlap, OMTG~\cite{xu2026omtg} supports multiple disjoint segments, and TimeLens~\cite{zhang2026timelens} combines re-annotated data, textual time encoding, and thinking-free reinforcement learning.
Spatial grounding focuses on object localization and correspondence across frames. VideoRefer~\cite{zhang2025videorefer} combines region-level instructions with a spatio-temporal object encoder, Sa2VA~\cite{yuan2025sa2va} connects an MLLM with SAM-2~\cite{ravi2025sam2} for temporally consistent masks, and SpaceR~\cite{ouyang2025spacer} constructs explicit maps of object layouts and relations. Joint spatio-temporal methods include STCAT~\cite{jin2022stcat}, which associates object-tube predictions with a consistency-aware transformer; LRR~\cite{bhattacharyya2023lrr}, which organizes grounded reasoning into look, remember, and reason stages; LLaVA-ST~\cite{li2025llava-st}, which aligns coordinate tokens while decoupling spatial and temporal compression; and STVG-o1~\cite{gu2025thinking}, which inserts bounding-box reasoning before localization. 
Our previous Open-o3-Video~\cite{meng2025open} places timestamps, object references, and boxes directly inside a reasoning trace, but produces them in one round and cannot acquire evidence omitted by the initial observation. Open-o3-Video-v2 extends this formulation into an iterative evidence-grounded process, where missing observations are actively acquired, grounded to supporting moments and objects, and incorporated into subsequent reasoning.

\noindent\textbf{Video Reasoning.}
Video reasoning methods can be broadly divided into text-only reasoning and reasoning with videos~\cite{meng2026survey}. Text-only approaches generate a rationale after the video has been encoded or converted to textual memory. They learn this behavior through chain-of-thought supervision~\cite{zhang2025video-cot,ghazanfari2025chain}, reinforcement learning~\cite{feng2025video-r1,wang2025videorft,wang2025video-rts,park2025deepvideo,dang2025twgrpo}, or preference optimization~\cite{huang2025vistadpo,li2025veripo}. Video-R1 encourages temporally aware reasoning, VideoRFT rewards semantic agreement between reasoning and video features, and DeepVideo-R1 stabilizes optimization through advantage regression and difficulty-aware data. Other systems organize captions, memories, or retrieval indexes for multi-stage inference and verification~\cite{fan2024videoagent,yang2024doraemongpt,fei2024video,ren2025videorag,wang2025videotree,zhi2025videoagent2,jin2025cot}. Although effective, their decisive intermediate states remain textual and do not expose precise visual support.

Thinking-with-video methods instead go beyond a fixed visual context by interleaving inference with explicit visual acquisition~\cite{openai-o3}. Agent-based methods differ in the granularity of their actions. VideoChat-R1.5~\cite{yan2025videochat} scales iterative perception; VITAL~\cite{zhang2025thinking} and FrameMind~\cite{ge2025framemind} request task-relevant clips or frames; and Pixel Reasoner~\cite{wang2025pixel} adds frame selection and spatial zoom. For long-video search, VideoExplorer~\cite{yuan2025videoexplorerthinkvideosagentic} and Conan~\cite{ouyang2025conan} formulate subquestions or identify evidence frames, while LongVT~\cite{yang2025longvt} and VideoZoomer~\cite{ding2025videozoomer} use temporal cropping and denser resampling in a global-to-local loop. At the policy level, Video-o3~\cite{zeng2026video} learns multi-hop clue seeking with adaptive termination, VideoTemp-o3~\cite{liu2026videotemp} couples tool use with temporal localization and correction, and VideoSeeker~\cite{zhao2026videoseeker} extends native tool invocation to instance-level spatio-temporal perception.
Evidence-based approaches require the reasoning trace to identify supporting observations through timestamps, dense captions, boxes, or other visual references~\cite{zhang2025vtimecot,wang2025video-thinker,zhang2025rewatch,maaz2025videor2}. Open-o3-Video~\cite{meng2025open} directly generates timestamped object boxes but remains restricted to the initial sample. Most agent-based methods can retrieve additional frames or clips, but rarely link the retrieved observations to spatio-temporal evidence supporting reasoning steps. Open-o3-Video-v2 addresses this gap with an agentic evidence-grounded formulation. Temporal calls recover observations missing from the initial context, while timestamped object boxes link reasoning to identifiable visual referents and make the acquired evidence human-verifiable.

\section{Open-o3-Video}
\label{sec:open_o3_video_v1}


Reliable video reasoning requires the model to identify the visual evidence that supports its answer.
\textit{Open-o3-Video} addresses this problem by embedding explicit temporal and object-level evidence into the reasoning trace. Given a video question, the model predicts the answer while identifying when the relevant event occurs and where the supporting object appears.
In this section, we briefly review the STGR data, grounded-thinking objective, and remaining limitations of Open-o3-Video that motivate \textit{Open-o3-Video-v2}.

\begin{figure*}[t]
    \centering
    \includegraphics[width=1.0\linewidth]{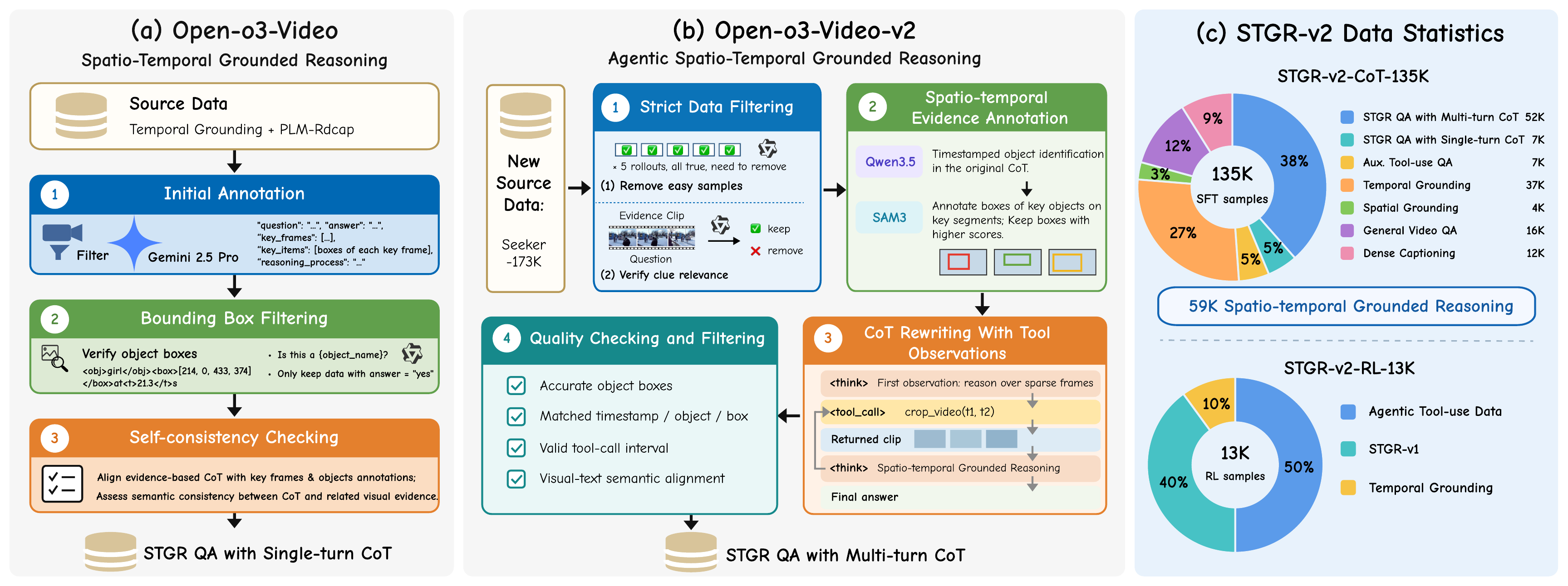}
    \caption{Overview of the data construction pipelines and STGR-v2 data statistics. \textbf{(a)} Open-o3-Video constructs single-turn spatio-temporal grounded reasoning data through initial annotation, bounding-box filtering, and self-consistency checking. \textbf{(b)} Open-o3-Video-v2 extends this pipeline to agentic multi-turn reasoning through strict data filtering, spatio-temporal evidence annotation, CoT (chain-of-thought) rewriting with tool observations, and quality checking. \textbf{(c)} STGR-v2-CoT-135K contains 135K SFT samples across seven task groups, including 59K spatio-temporal grounded reasoning samples, while STGR-v2-RL-13K contains 13K RL samples selected through strict filtering.}
    \label{fig:data_construction_v2}
\end{figure*}

\subsection{STGR Data Construction}
\label{sec:v1_data}

Open-o3-Video constructs STGR-CoT-30K for supervised fine-tuning (SFT) and STGR-RL-36K for reinforcement learning (RL), as summarized in Fig.~\ref{fig:data_construction_v2}(a). Both corpora mix temporal grounding, spatial grounding, spatio-temporal reasoning, and general video QA. Each grounded trace links a reasoning claim to a timestamped frame, an object name, and a bounding box in a parseable format. We filter visually inconsistent annotations so that intermediate evidence remains directly evaluable. The complete data composition, annotation format, and quality-control pipeline are reported in Appendix~A.

\begin{figure*}[t]
    \centering
    \includegraphics[width=1.0\linewidth]{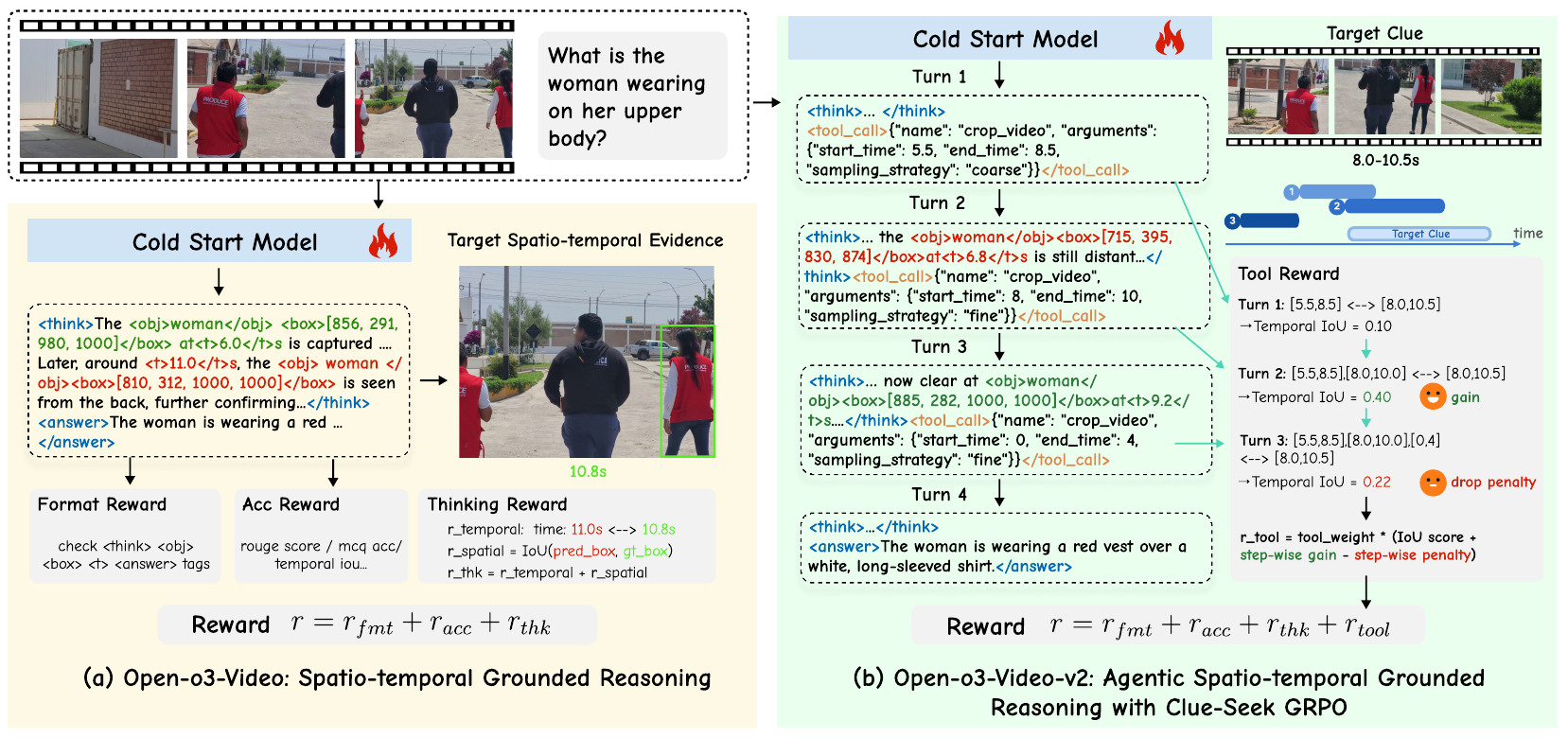}
    \caption{Overview of the reinforcement learning designs in Open-o3-Video and Open-o3-Video-v2. \textbf{(a)} Open-o3-Video produces a single-turn spatio-temporal grounded reasoning trace and optimizes it with format, answer-accuracy, and grounded thinking rewards; the thinking reward jointly evaluates temporal proximity and spatial overlap with the target evidence. \textbf{(b)} Open-o3-Video-v2 extends this formulation to multi-turn agentic reasoning with CS-GRPO. The model iteratively calls \texttt{crop\_video} to acquire denser observations, while CS-GRPO tracks how the accumulated evidence evolves across calls and assigns credit according to whether each interaction contributes useful new evidence.}
    \label{fig:rl_framework_v2}
\end{figure*}

\subsection{Training with Grounded Thinking Rewards}
\label{sec:v1_training}

We train Open-o3-Video in two stages. SFT teaches the evidence-grounded output pattern through responses containing \texttt{<think>} and \texttt{<answer>} blocks. In the subsequent RL stage, each response is scored by answer, thinking, and format rewards, as illustrated in Fig.~\ref{fig:rl_framework_v2}(a):
\begin{equation}
R(x,y)
=r_{\mathrm{acc}}(x,y)
+r_{\mathrm{thk}}(x,y)
+r_{\mathrm{fmt}}(x,y).
\end{equation}
Here, $r_{\mathrm{acc}}$ evaluates final-answer correctness and $r_{\mathrm{fmt}}$ checks the required output format. The grounded-thinking reward evaluates evidence cited in \texttt{<think>} and is decomposed as
\begin{equation}
r_{\mathrm{thk}}(x,y)
=r_{\mathrm{t}}(x,y)
+r_{\mathrm{s}}(x,y).
\end{equation}
The temporal term $r_{\mathrm{t}}$ evaluates whether cited timestamps agree with interval or point supervision. For point supervision, \textbf{Adaptive Temporal Proximity} anneals the temporal neighborhood from broad to strict, providing smoother feedback early in training while preserving precise localization later.
The spatial term $r_{\mathrm{s}}$ evaluates the IoU between a predicted object box and the annotation at its matched timestamp. \textbf{Temporal Gating} activates this spatial reward only when the predicted timestamp is sufficiently close to the annotated time, preventing boxes at irrelevant moments from receiving credit. The complete reward definitions are provided in Appendix A.

\subsection{Remaining Challenges in Open-o3-Video}
\label{sec:v1_limitation}


Open-o3-Video still has two limitations in complex video scenarios.
\textbf{First}, it reasons within a fixed visual context and cannot request denser observations when sparse sampling misses a critical clue.
\textbf{Second}, its grounded reasoning supervision remains limited in scale and diversity.
Recent agentic video methods overcome fixed observations through temporal zooming~\cite{zeng2026video,ding2025videozoomer,zhao2026videoseeker}, but retrieving an interval mainly determines where to observe and does not by itself identify which moment and visual entity support a reasoning step.
These limitations motivate Open-o3-Video-v2, which extends the fixed-context formulation into an active process of acquiring, grounding, and integrating missing visual evidence.

\section{Open-o3-Video-v2}
\label{sec:open_o3_video_v2}


\begin{sloppypar}
Building on the Open-o3-Video formulation reviewed in Sec.~\ref{sec:open_o3_video_v1}, Open-o3-Video-v2 extends evidence-grounded reasoning from a fixed observation setting to an active process of acquiring, grounding, and integrating missing visual evidence.
The model can identify unresolved information during reasoning, selectively inspect a relevant temporal interval, and convert the returned observation into explicit spatio-temporal evidence for subsequent inference.
This section presents the new v2 formulation in detail, including the multi-turn reasoning process, STGR-v2 data pipeline, and Clue-Seek GRPO (CS-GRPO) for optimizing evidence acquisition over interaction trajectories.
\end{sloppypar}

\subsection{Task Formulation}
\label{sec:v2_task_formulation}

We represent a video of duration $T$ as
\begin{equation}
V=\{(t_i,I_i)\}_{i=1}^{N},
\qquad 0\leq t_1<\cdots<t_N\leq T,
\end{equation}
where $N$ is the number of frames available to the observation system and $I_i$ is the frame at timestamp $t_i$ (in seconds).
Given a question $q$, the default sparse sampler $\mathcal{S}_0$ produces the initial observation $O_0=\mathcal{S}_0(V)$, yielding the model input $x=(q,O_0)$.
The policy generates an interleaved trajectory containing $K\geq0$ tool calls,
\begin{equation}
\tau=(z_1,o_1,z_2,o_2,\ldots,z_K,o_K,z_{K+1}).
\end{equation}
For $k\leq K$, $z_k$ contains a \texttt{<think>} block followed by a \texttt{<tool\_call>} block, and $o_k$ is the returned observation; the terminal response $z_{K+1}$ instead ends with \texttt{<answer>}$a$\texttt{</answer>}, where $a$ is the final answer.

\noindent\textbf{Tool call.}
At non-final turn $k$, the model requests a segment using
\begin{equation}
u_k=(s_k,e_k,d_k),
\end{equation}
where $0\leq s_k<e_k\leq T$ bounds the interval and $d_k\in\{\mathrm{coarse},\mathrm{medium},\mathrm{fine}\}$ controls its temporal sampling density.
Similar to prior agentic video reasoning methods~\cite{zeng2026video}, the request is serialized as a \texttt{<tool\_call>} to \texttt{crop\_video} with \texttt{start\_time}, \texttt{end\_time}, and \texttt{sampling\_strategy}.
The tool returns $o_k=\mathcal{C}(V,u_k)$, comprising frames sampled from $[s_k,e_k]$ according to $d_k$, and appends it to the context.
Each response is therefore conditioned on the question, initial observation, and preceding interaction history:
\begin{equation}
z_k \sim \pi_{\theta}(\cdot \mid q,O_0,z_{<k},o_{<k}),
\qquad k=1,\ldots,K+1,
\end{equation}
where $z_{<k}=(z_1,\ldots,z_{k-1})$ and $o_{<k}=(o_1,\ldots,o_{k-1})$.

\noindent\textbf{Embedded evidence.}
The \texttt{<think>} blocks may also contain explicit spatio-temporal evidence claims, each parsed as
\begin{equation}
\varepsilon_m=(\hat{t}_m,c_m,b_m),
\end{equation}
where $\hat{t}_m\in[0,T]$ is the claimed timestamp in seconds, $c_m$ is the object name, and $b_m=[x_{1m},y_{1m},x_{2m},y_{2m}]$ is its bounding box.
The two coordinate pairs give the top-left and bottom-right corners, respectively, in the model's $1000\times1000$ image coordinate system.
We consistently order evidence as timestamp, object name, and bounding box, and collect all $M$ valid claims in $\mathcal{E}(\tau)=\{\varepsilon_m\}_{m=1}^{M}$.
The resulting evidence claims remain in the interaction history and can serve as explicit premises for subsequent reasoning. Their structured form also makes the intermediate evidence directly measurable during optimization.

\subsection{STGR-v2: Data Optimization}
\label{sec:v2_data}

\noindent\textbf{From grounded responses to agentic trajectories.}
Open-o3-Video-v2 represents samples in a conversation format that supports both single-turn responses and multi-turn interactions. For tool-use trajectories, this format interleaves reasoning, tool calls, returned frames, embedded object-level evidence, and a final answer. Compared with v1's single response, these trajectories jointly teach when to inspect a clip, which evidence within the returned observation supports the reasoning, and how that evidence should be integrated into later inference. We additionally retain single-turn grounding, temporal localization, and dense-caption supervision to preserve complementary video capabilities.

\noindent\textbf{Agentic data construction.}
We derive 52K spatio-temporal grounded tool-use QA trajectories from Seeker-173K~\cite{zeng2026video} through the four stages in Fig.~\ref{fig:data_construction_v2}(b).

\noindent\textbf{(1) Strict filtering.}
Qwen3-VL~\cite{bai2025qwen3vl} performs five rollouts from eight sparse frames; we remove questions answered correctly in every rollout because they do not require another observation. We also discard samples that remain unanswerable when the annotated clue clip is supplied, ensuring that retained clues support the answer.

\noindent\textbf{(2) Evidence annotation.}
Qwen3.5~\cite{qwen3.5} extracts question-relevant objects and their temporal spans; SAM3~\cite{carion2025sam3} then produces boxes on sampled clue frames. We remove boxes covering more than 80\% of a frame and retain the two highest-confidence boxes per object and target segment.

\noindent\textbf{(3) Agentic CoT rewriting.}
Gemini-2.5-Pro~\cite{comanici2025gemini} rewrites the original rationale using the verified clue and boxes. The resulting trajectory contains sparse-observation reasoning, tool calls, returned frames, grounded reasoning in the v1 evidence format, and a final answer.

\noindent\textbf{(4) Quality checking and filtering.}
We apply four checks to the constructed trajectories. 
First, each bounding box should accurately localize the target object without being excessively large or imprecise. Second, the cited timestamp, object, and bounding box should be correctly matched. For these two checks, we first filter boxes exceeding an area threshold and then crop the corresponding region from the cited frame for Qwen3.5~\cite{qwen3.5} to verify whether it correctly localizes the referenced object instance.
Third, we apply rule-based checks to ensure that each tool call specifies a valid temporal interval within the video range.
Finally, we use Qwen3.5 to verify whether each reasoning description is semantically supported by its cited visual evidence.
Samples that fail these checks are removed unless the inconsistency can be resolved unambiguously.

We also manually audit 500 randomly sampled examples covering different sources and task types after stages~(2) and~(3). The audits examine box quality, annotation consistency, tool-call validity, and the reasoning logic. More than 90\% of the audited examples satisfy the corresponding quality criteria at each stage, indicating that the annotation and trajectory-construction pipeline is generally reliable. Recurring errors identified during the audits are incorporated into the checking and filtering criteria in stage~(4). Detailed error categories and inter-annotator agreement are reported in Appendix B.

\noindent\textbf{SFT corpus.}
\textbf{STGR-v2-CoT-135K} contains 52K multi-turn and 7K single-turn spatio-temporal grounded QA samples, 7K auxiliary tool-use samples, 37K temporal- and 4K spatial-grounding samples, 16K general video QA samples, and 12K dense captions (Fig.~\ref{fig:data_construction_v2}(c)). Sources include Seeker-173K, STGR-CoT-30K, CG-Bench, OMTG, TimeLens, OmniVideo-100K, and CapRL-Video-178K~\cite{zeng2026video,meng2025open,CGBench,xu2026omtg,zhang2026timelens,cai2026omnivideo100k,yang2026caprlplusplus}. Thus, 59K samples (44\%) provide direct spatio-temporal grounded QA supervision, while the remaining tasks preserve localization and general video perception.

\noindent\textbf{RL corpus.}
\textbf{STGR-v2-RL-13K} combines 40\% v1 data, 50\% agentic tool-use data, and 10\% TimeLens temporal grounding~\cite{zhang2026timelens}. From five SFT-policy responses sampled at temperature 1.0, we retain questions with both correct and incorrect rollouts, removing already-solved and currently intractable prompts to provide informative reward contrast.

\subsection{Clue-Seek GRPO}
\label{sec:cs_grpo}

Multi-turn observation is useful only when the model learns how each newly acquired observation changes what is already known. An effective policy must identify missing evidence, inspect a relevant interval, integrate the returned observation with the accumulated context, and stop once sufficient support has been obtained. Standard GRPO with task-level supervision can optimize the final outcome, but provides limited information about which intermediate observation actually improves the evidence available for reasoning.

We introduce Clue-Seek GRPO (CS-GRPO) to optimize this sequential evidence-acquisition process. As illustrated in Fig.~\ref{fig:rl_framework_v2}(b), CS-GRPO maintains an evolving evidence state over the interaction history and uses its transitions to evaluate successive tool interactions. This makes the utility of a call history-dependent: a retrieved interval is valuable when it contributes new answer-relevant evidence, but less useful when it repeatedly covers information that has already been observed.
The resulting optimization encourages the policy to seek informative intervals, accumulate complementary evidence, and terminate when further observation provides little additional value.

For a complete trajectory $\tau$, let $\{p_k=[s_k,e_k]\}_{k=1}^{K}$ be the intervals from its $K$ valid calls.
Let $\mathcal{G}=\{g_j\}_{j=1}^{J}$ denote the set of $J$ annotated answer-relevant clue intervals.
Let $N_{\mathrm{call}}$ count all complete \texttt{<tool\_call>} blocks, including malformed calls excluded from the $K$ valid intervals.
The evidence state after valid call $k$ is the multi-segment temporal IoU between all intervals retrieved so far and the annotated clues,
\begin{equation}
\gamma_k=
\frac{\left|\left(\cup_{l=1}^{k}p_l\right)\cap\left(\cup_{j=1}^{J}g_j\right)\right|}
{\left|\left(\cup_{l=1}^{k}p_l\right)\cup\left(\cup_{j=1}^{J}g_j\right)\right|},
\end{equation}
where $|\cdot|$ measures temporal duration. We set $\gamma_0=0$, so a trajectory without a valid call has zero clue coverage.
The scalar $\gamma_k$ jointly measures clue recall and irrelevant temporal coverage accumulated up to step $k$.

Rather than scoring each call in isolation, we use the sequence $\{\gamma_k\}_{k=0}^{K}$ to characterize how the evidence available to the model evolves throughout the trajectory. CS-GRPO rewards improvements in this state while regularizing acquisition patterns that add little useful evidence.

Two terms characterize final evidence quality and step-wise progress.
The terminal clue reward
\begin{equation}
r_{\mathrm{clue}}(\tau)=\gamma_K
\end{equation}
evaluates final clue coverage while accounting for irrelevant temporal content. Because it does not identify which call improved the state, we also define
\begin{equation}
r_{\mathrm{gain}}(\tau)=
\sum_{k=2}^{K}
\mathbb{I}\left[\gamma_k-\gamma_{k-1}>0.1\right],
\end{equation}
where $\mathbb{I}[\cdot]$ is the indicator function.
This rewards a subsequent call when it adds sufficient clue-aligned evidence, and the first call remains evaluated by $r_{\mathrm{clue}}$.

Evidence-state improvement alone does not discourage inefficient acquisition patterns, so we additionally regularize excessive, repeated, fragmented, and degrading calls.
Because annotated clues can typically be found within three calls, the budget penalty
\begin{equation}
r_{\mathrm{budget}}(\tau)=
\max(0,N_{\mathrm{call}}-3)
\end{equation}
counts calls beyond this allowance.

The repetition penalty discourages nearly identical intervals:
\begin{equation}
r_{\mathrm{repeat}}(\tau)=
\sum_{1\leq i<j\leq K}
\mathbb{I}\left[\mathrm{tIoU}(p_i,p_j)>0.7\right],
\end{equation}
where overlapping pairs with temporal IoU above $0.7$ are counted.

To discourage sequential scanning through adjacent fragments rather than selecting promising clues, we introduce a fragmentation penalty.
For consecutive intervals $p_i=[s_i,e_i]$ and $p_{i+1}=[s_{i+1},e_{i+1}]$, we treat them as adjacent when $-10^{-6}\leq s_{i+1}-e_i\leq10^{-6}$.
Each maximal sequence of at least three consecutive adjacent intervals contributes one count to $r_{\mathrm{frag}}(\tau)$.

Finally, the drop penalty discourages a subsequent call from diluting already located evidence:
\begin{equation}
r_{\mathrm{drop}}(\tau)=
\sum_{k=2}^{K}
\mathbb{I}\left[\gamma_{k-1}-\gamma_k>0.1\right],
\end{equation}
which counts state decreases larger than $0.1$.

We summarize these evidence-state transitions into a trajectory-level tool-use score:
\begin{equation}
\begin{aligned}
r_{\mathrm{tool}}(x,\tau)={}&\mathrm{clip}_{[0,1]}\big(
r_{\mathrm{clue}}(\tau)+0.1r_{\mathrm{gain}}(\tau)-0.1r_{\mathrm{budget}}(\tau)\\
&-0.1r_{\mathrm{repeat}}(\tau)-0.1r_{\mathrm{frag}}(\tau)-0.05r_{\mathrm{drop}}(\tau)\big).
\end{aligned}
\end{equation}
This score provides trajectory-level supervision for the evolution of acquired evidence while retaining the standard GRPO estimator. CS-GRPO thereby makes three coupled decisions directly optimizable: where to search, whether a new observation improves the accumulated evidence, and when further observation is no longer beneficial.

\subsection{Reward Design}
\label{sec:v2_reward}

The final objective jointly evaluates the quality of the answer, grounded reasoning, trajectory structure, and evidence acquisition.
For model input $x$ and trajectory $\tau$, we first clip each component to $[0,1]$ and compute
\begin{equation}
R(x,\tau)=w_{\mathrm{acc}}r_{\mathrm{acc}}(x,\tau)+w_{\mathrm{fmt}}r_{\mathrm{fmt}}(x,\tau)+w_{\mathrm{thk}}r_{\mathrm{thk}}(x,\tau)+w_{\mathrm{tool}}r_{\mathrm{tool}}(x,\tau),
\label{eq:v2_reward}
\end{equation}
where $w_{\mathrm{acc}}$, $w_{\mathrm{fmt}}$, $w_{\mathrm{thk}}$, and $w_{\mathrm{tool}}$ are nonnegative coefficients that control the contribution of each component.

\noindent
\textit{The answer reward} $r_{\mathrm{acc}}(x,\tau)$ follows the task-specific metrics used in Open-o3-Video, including option matching, text similarity, spatial IoU, and temporal IoU.

\noindent
\textit{The binary format reward} $r_{\mathrm{fmt}}(x,\tau)$ equals one when the complete trajectory satisfies all output requirements and otherwise equals zero.
Its major tags must follow
\begin{center}
\small
\begin{tabular}{l}
\texttt{(<think></think><tool\_call></tool\_call>)*}\\
\texttt{<think></think><answer></answer>}
\end{tabular}
\end{center}
The major tags must also be balanced and non-crossing.
The minor tags \texttt{<obj>}, \texttt{<box>}, and \texttt{<t>} must be balanced.
Each tool call must be valid JSON, invoke \texttt{crop\_video}, use a legal sampling strategy, and satisfy $e_k>s_k$.
For temporal-answer tasks, the answer must follow \texttt{From <t>start</t>s to <t>end</t>s}.

\noindent
\textit{The thinking reward} $r_{\mathrm{thk}}(x,\tau)$ extends the grounded thinking reward in Sec.~\ref{sec:v1_training} from a single response to all \texttt{<think>} blocks in the trajectory.
It evaluates the timestamps, objects, and boxes explicitly cited in reasoning using the available spatio-temporal annotations, and therefore provides the primary supervision for object-level grounding correctness.

\begin{sloppypar}
\noindent
\textit{The tool reward} $r_{\mathrm{tool}}(x,\tau)$ follows the temporal evidence-acquisition score defined in Sec.~\ref{sec:cs_grpo} for clue-supervised samples.
For samples without clue intervals, it provides lightweight regularization against excessive, repeated, or fragmented calls.
The two terms optimize complementary aspects of agentic evidence-grounded reasoning: $r_{\mathrm{tool}}$ mainly supervises \emph{where to observe}, while $r_{\mathrm{thk}}$ supervises \emph{how the acquired observation is grounded} to fine-grained moments and objects. They are jointly optimized in Eq.~\ref{eq:v2_reward}.
\end{sloppypar}

\subsection{Training Recipes}
\label{sec:v2_training_recipes}

The v2 model is trained with SFT followed by CS-GRPO. SFT represents each sample in a conversation format that supports both single-turn responses and multi-turn interactions, with user video context, assistant reasoning, optional tool calls and returned frames, grounded evidence, and a final answer. The standard next-token loss is applied only to assistant-generated tokens; user prompts, videos, and tool observations are masked as context.
CS-GRPO then optimizes complete trajectories from the agent loop using the objective in Sec.~\ref{sec:v2_reward}, where evidence-state transitions provide additional supervision for successive observation decisions.
%

\subsection{Training Infrastructure}
\label{sec:v2_infra}

The static, task-specific v1 trainers do not naturally support long trajectories that alternate between model actions and returned video observations. We therefore perform SFT with \textbf{ms-swift\footnote{\url{https://github.com/modelscope/ms-swift}}}, whose multimodal conversation format and distributed backends support scalable training on the mixed 135K-sample corpus, and implement \texttt{crop\_video} with the \textbf{verl\footnote{\url{https://github.com/verl-project/verl}}} AgentLoop for RL. The loop executes tools between model turns, masks returned frames as context, retains model tokens as policy actions, and passes each complete trajectory to the objective in Sec.~\ref{sec:v2_reward}. We have open-sourced this training stack.

\section{Experiments}
\label{sec:exp}

\begin{table*}[t]
  \centering
  \caption{Performance on V-STaR (\%). Chain1 denotes \textit{what--when--where}, whereas Chain2 denotes \textit{what--where--when}. mAM and mLGM jointly summarize answer accuracy and spatio-temporal alignment. Bold and underlined numbers denote the best and second-best results, respectively.}
  \label{tab:vstar}
  \setlength{\tabcolsep}{4pt}
  \scriptsize
  \begin{tabular}{lccccccc}
    \toprule[0.15em]
    \textbf{Model} &
    \textbf{What} &
    \multicolumn{2}{c}{\textbf{When (Temporal IoU)}} &
    \multicolumn{2}{c}{\textbf{Where (Visual IoU)}} &
    \multicolumn{2}{c}{\textbf{Overall}} \\
    \cmidrule(lr){2-2} \cmidrule(lr){3-4} \cmidrule(lr){5-6} \cmidrule(lr){7-8}
     & \textbf{Acc} & \textbf{Chain1} & \textbf{Chain2}
     & \textbf{Chain1} & \textbf{Chain2} & \textbf{mAM} & \textbf{mLGM} \\
    \midrule
    GPT-4o~\cite{hurst2024gpt4o} & 60.8 & 16.7 & 12.8 & 6.5 & 3.0 & 26.8 & 38.2 \\
    Gemini-2-Flash~\cite{comanici2025gemini} & 53.0 & 24.5 & 23.8 & 4.6 & 2.2 & 26.9 & 35.6 \\
    Gemini-3-Pro~\cite{comanici2025gemini} & 59.1 & 24.2 & 23.8 & 7.2 & 4.8 & 29.9 & 41.2 \\
    \midrule
    Video-LLaMA3-7B~\cite{zhang2025videollama3} & 41.9 & 23.0 & 23.1 & 0.9 & 0.2 & 21.7 & 27.0 \\
    LLaVA-Video-7B~\cite{zhang2024llavavideo} & 49.5 & 10.5 & 12.2 & 1.9 & 1.3 & 20.8 & 27.3 \\
    VideoChat2-7B~\cite{videochat2} & 36.2 & 13.7 & 12.5 & 2.5 & 1.0 & 17.0 & 20.3 \\
    InternVL-2.5-8B~\cite{internvl2.5} & 44.2 & 8.7 & 7.8 & 0.7 & 0.1 & 17.6 & 24.9 \\
    Qwen2.5-VL-7B~\cite{qwen2.5vl} & 33.5 & 15.4 & 13.8 & 17.0 & 2.5 & 19.3 & 22.4 \\
    VideoZoomer-7B~\cite{ding2025videozoomer} & 56.5 & 20.4 & 20.8 & 2.8 & 0.7 & 26.2 & 36.0 \\
    VideoTemp-o3-7B~\cite{liu2026videotemp} & 37.0 & 16.8 & 24.1 & 1.3 & 0.4 & 19.4 & 23.4 \\
    Video-o3-7B~\cite{zeng2026video} & 31.7 & 26.8 & 25.2 & 0.0 & 0.0 & 19.3 & 22.8 \\
    Qwen3-VL-4B~\cite{bai2025qwen3vl} & 59.6 & 18.6 & 19.2 & 26.8 & 7.7 & 31.9 & 43.7 \\
    Qwen3-VL-8B~\cite{bai2025qwen3vl} & 42.9 & 23.5 & 24.4 & 27.3 & 7.1 & 28.0 & 34.4 \\
    \midrule
    TRACE~\cite{guo2025trace} & 17.6 & 19.1 & 17.1 & 0.0 & 0.0 & 12.0 & 13.3 \\
    Sa2VA-8B~\cite{yuan2025sa2va} & 16.4 & 0.1 & 0.0 & 32.3 & \textbf{37.5} & 17.1 & 20.3 \\
    \midrule
    Open-o3-Video-7B~\cite{meng2025open} & 61.0 & 24.5 & 24.0 & 25.4 & 6.0 & 33.7 {\scriptsize(+14.4\%)} & 46.6 {\scriptsize(+24.2\%)} \\
    Open-o3-Video-4B & 59.7 & 25.1 & 23.8 & 31.2 & 8.1 & 34.6 {\scriptsize(+2.7\%)} & 47.3 {\scriptsize(+3.6\%)} \\
    Open-o3-Video-8B & 61.1 & 24.9 & 23.5 & 33.2 & 8.8 & 35.5 {\scriptsize(+7.5\%)} & 49.0 {\scriptsize(+14.6\%)} \\
    \midrule
    Open-o3-Video-v2-4B & \textbf{62.8}  & \textbf{28.8}  & \textbf{30.0}  & \underline{38.8} & 19.1 & \underline{40.4} {\scriptsize(+8.5\%)} & \underline{56.3} {\scriptsize(+12.6\%)} \\
    Open-o3-Video-v2-8B & \underline{62.5}  & \underline{27.9}  & \underline{28.0}  & \textbf{39.5} & \underline{23.9} & \textbf{40.7} {\scriptsize(+12.7\%)} & \textbf{56.5} {\scriptsize(+22.1\%)} \\
    \bottomrule[0.15em]
  \end{tabular}
\end{table*}

\noindent
\textbf{Benchmarks and evaluation metrics.}
We evaluate three video understanding tasks, including \textit{spatio-temporal grounded question answering}, \textit{general video understanding}, and \textit{temporal grounding}.
We use a different set of benchmarks for each task.
For spatio-temporal grounded question answering, V-STaR~\cite{VSTaR} measures three complementary capabilities: answering the question (\textit{What}), localizing the supporting moment (\textit{When}), and grounding the relevant object (\textit{Where}).
It evaluates two reasoning orders, \textit{what--when--where} (Chain1) and \textit{what--where--when} (Chain2), and reports answer accuracy, temporal IoU, visual IoU, and the composite mAM and mLGM scores.
For general video understanding, we use VideoMME~\cite{VideoMME}, VideoMME-v2~\cite{VideoMMEv2}, WorldSense~\cite{WorldSense}, VideoMMMU~\cite{VideoMMMU}, Video-Holmes~\cite{cheng2025videoholmes}, and LVBench~\cite{LVBench}.
Together, these benchmarks cover broad video comprehension, real-world multimodal perception, knowledge-intensive reasoning, multi-hop clue inference, and extremely long videos.
For temporal grounding, we evaluate on Charades-TimeLens~\cite{zhang2026timelens} and TVGBench~\cite{TVGBench}, both of which require a model to localize the temporal interval described by a language query; we report mean temporal IoU (mIoU).

\noindent
\textbf{Implementation details.}
Open-o3-Video is instantiated with Qwen2.5-VL-7B~\cite{qwen2.5vl} and, for controlled comparisons with v2, we also train it from Qwen3-VL-4B and Qwen3-VL-8B. Open-o3-Video-v2 uses the same two Qwen3-VL backbones~\cite{bai2025qwen3vl}.
Both generations are trained on NVIDIA H100 GPUs.
For Open-o3-Video, we uniformly sample 16 frames from each video, cap the resolution of each frame at $128 \times 28 \times 28$, and insert annotated key frames when they are available.
Each frame is prepended with its absolute timestamp to strengthen temporal perception.
For Open-o3-Video-v2, SFT uniformly samples the full video at a target rate of 1 FPS with up to 96 frames and uses coarse-to-fine cropped observations for multi-turn samples.
During CS-GRPO, each rollout starts from at most 32 uniformly distributed frames, and the three crop strategies progressively increase the number of frames returned from the requested interval.
At inference time, the initial observation is sampled at 1 FPS with at most 96 frames, and each trajectory is limited to at most five assistant turns.
We perform full-parameter SFT and CS-GRPO for one epoch each, using learning rates of $5\times10^{-6}$ and $1\times10^{-6}$, respectively. CS-GRPO is trained on 13K samples with eight rollouts per prompt.
More detailed training and evaluation configurations are provided in Appendix~\ref{app-sec:implementation_details}.

\subsection{Main Results}
\label{sec:main_results}

\begin{table*}[t]
  \centering
  \caption{Performance on general video understanding and temporal grounding benchmarks (\%). For WorldSense, we report overall and recognition accuracy. The last two columns report mIoU. All other columns report accuracy. Bold and underlined numbers denote the best and second-best results, respectively.}
  \label{tab:general_and_temporal}
  \setlength{\tabcolsep}{1.8pt}
  \scriptsize
  \begin{tabular}{lccccccccc}
    \toprule[0.15em]
    \textbf{Model} &
    \multicolumn{7}{c}{\textbf{General Video Understanding and Reasoning}} &
    \multicolumn{2}{c}{\textbf{Temporal Grounding}} \\
    \cmidrule(lr){2-8} \cmidrule(lr){9-10}
     & \makecell{\textbf{Video-}\\\textbf{Holmes}} & \makecell{\textbf{LV}\\\textbf{Bench}}
     & \makecell{\textbf{Video-}\\\textbf{MMMU}}
     & \multicolumn{2}{c}{\textbf{WorldSense}}
     & \makecell{\textbf{Video}\\\textbf{MME}} & \makecell{\textbf{Video}\\\textbf{MME-v2}}
     & \makecell{\textbf{Charades-}\\\textbf{TimeLens}} & \makecell{\textbf{TVG}\\\textbf{Bench}} \\
    \cmidrule(lr){5-6}
     & \textbf{Acc} & \textbf{Acc} & \textbf{Acc}
     & \textbf{Overall} & \textbf{Recog.}
     & \textbf{Acc} & \textbf{Acc} & \textbf{mIoU} & \textbf{mIoU} \\
    \midrule
    Qwen2.5-VL-7B~\cite{qwen2.5vl}
    & 34.7 & 45.3 & 51.2 & 36.1 & 33.7 & 62.4 & 25.3 & 39.3 & 16.3 \\

    Qwen3-VL-4B~\cite{bai2025qwen3vl}
    & 37.3 & 46.0 & 54.4 & 41.7 & 42.0 & 65.3 & 26.2 & 45.6 & 27.1 \\

    Qwen3-VL-8B~\cite{bai2025qwen3vl}
    & 44.2 & 47.3 & 57.3 & \underline{45.7} & 44.0
    & \underline{67.4} & \underline{27.9} & 48.3 & 29.4 \\

    VideoRFT-7B~\cite{wang2025videorft}
    & 40.3 & 40.4 & 51.1 & 38.2 & 36.6 & 59.8 & 22.4 & 17.0 & 14.3 \\

    VideoR1-7B~\cite{feng2025video-r1}
    & 42.6 & 39.8 & 52.4 & 35.5 & 32.8 & 61.4 & 22.7 & 20.8 & 9.6 \\

    VideoZoomer-7B~\cite{ding2025videozoomer}
    & 45.7 & 41.5 & 52.2 & 42.4 & 43.1 & 65.2 & 23.6 & 33.5 & 20.0 \\

    VideoTemp-o3-7B~\cite{liu2026videotemp}
    & 40.3 & 43.0 & 53.2 & 40.6 & 40.6 & 64.5 & 23.3 & 38.9 & 30.4 \\

    Video-o3-7B~\cite{zeng2026video}
    & \underline{46.5} & 47.6 & 51.7 & 45.4 & \underline{48.1}
    & 66.5 & 24.1 & 36.9 & 27.6 \\

    \midrule

    Open-o3-Video-7B~\cite{meng2025open}
    & 40.6 & 41.2 & 52.3 & 37.5 & 36.8 & 63.6 & 22.4 & 30.9 & 20.8 \\

    Open-o3-Video-v2-4B
    & \textbf{48.0} & \underline{53.2} & \underline{59.8}
    & \textbf{46.3} & \textbf{48.3}
    & 66.6 & 27.0 & \underline{49.8} & \textbf{31.3} \\

    Open-o3-Video-v2-8B
    & 46.1 & \textbf{53.9} & \textbf{63.0}
    & 45.5 & 47.2
    & \textbf{67.9} & \textbf{29.2} & \textbf{51.1} & \underline{30.9} \\

    \bottomrule[0.15em]
  \end{tabular}
\end{table*}

\noindent
\textbf{Spatio-temporal grounded QA on V-STaR.}
Table~\ref{tab:vstar} compares three complementary baseline families: frontier proprietary models (GPT-4o~\cite{hurst2024gpt4o} and Gemini~\cite{comanici2025gemini}), general-purpose open-source video MLLMs~\cite{zhang2025videollama3,zhang2024llavavideo,qwen2.5vl,zeng2026video,bai2025qwen3vl}, and task-specialized grounding methods (TRACE~\cite{guo2025trace} and Sa2VA~\cite{yuan2025sa2va}).
The composite mAM and mLGM metrics test whether a model can answer correctly while maintaining temporally and spatially consistent evidence.

\noindent
\textit{Open-o3-Video.}
Starting from Qwen2.5-VL-7B, Open-o3-Video-7B improves answer accuracy by 27.5\%, temporal IoU by 9.1\%/10.2\% on Chain1/Chain2, and visual IoU by 8.4\%/3.5\%.
Together, these gains increase mAM and mLGM by 14.4\% and 24.2\%, respectively, exceeding Gemini-3-Pro by 3.8\% and 5.4\% and achieving the strongest composite grounding performance among the compared contemporaneous baselines.
The improvement therefore covers not only answer accuracy but also alignment with explicit spatio-temporal evidence.
Even with limited dedicated temporal-localization supervision, Open-o3-Video-7B surpasses TRACE by 5.4\%/6.9\% temporal IoU on Chain1/Chain2, slightly exceeds Gemini-3-Pro by 0.3\%/0.2\%, and remains competitive with the agentic Video-o3 model.
For spatial grounding, Sa2VA retains the highest Chain2 visual IoU, but its weak answer and temporal performance limit its composite scores. Open-o3-Video instead substantially improves spatial IoU over Qwen2.5-VL and Video-o3 while preserving strong QA and temporal localization.
The Qwen3-VL experiments further confirm transfer across model scales: Open-o3-Video-4B improves its backbone by 2.7\% mAM and 3.6\% mLGM, while Open-o3-Video-8B gains 7.5\% and 14.6\%, respectively.
These consistently stronger composite scores verify the value of jointly optimizing answer correctness, temporal localization, and spatial grounding.

\noindent
\textit{Open-o3-Video-v2.}
The v2 models combine explicit evidence grounding with agentic, multi-turn clue acquisition.
Relative to Qwen3-VL-4B, Open-o3-Video-v2-4B improves answer accuracy by 3.2\%, temporal IoU by 10.2\%/10.8\%, and visual IoU by 12.0\%/11.4\%, resulting in gains of 8.5\% mAM and 12.6\% mLGM.
Open-o3-Video-v2-8B reaches the best overall results, 40.7\% mAM and 56.5\% mLGM, improving over Qwen3-VL-8B by 12.7\% and 22.1\%, respectively.
Compared with Open-o3-Video-4B, the v2 4B model improves answer accuracy by 3.1\%, temporal IoU by 3.7\%/6.2\%, visual IoU by 7.6\%/11.0\%, and mAM/mLGM by 5.8\%/9.0\%.
The 8B model similarly exceeds its v1 counterpart by 5.2\% mAM and 7.5\% mLGM.
These consistent gains show that combining evidence-grounded reasoning with agentic clue seeking enables more accurate spatio-temporal localization and stronger overall performance.

\noindent
\textbf{General video understanding and temporal grounding.}
Table~\ref{tab:general_and_temporal} evaluates generalization across a broader range of video understanding and grounding benchmarks.
We compare our model with Qwen baselines~\cite{qwen2.5vl,bai2025qwen3vl}, textual reasoning methods (Video-R1~\cite{feng2025video-r1}, VideoRFT~\cite{wang2025videorft}) and agentic clue-seeking methods~(VideoZoomer~\cite{ding2025videozoomer}, VideoTemp-o3~\cite{liu2026videotemp}, and Video-o3~\cite{zeng2026videoo3}).
Together, these benchmarks assess general video QA, long-video understanding, knowledge-intensive reasoning, multi-hop clue integration, and temporal grounding.

\noindent
\textit{Open-o3-Video.}
Relative to Qwen2.5-VL-7B, Open-o3-Video-7B improves VideoMME by 1.2\%, WorldSense overall/recognition by 1.4\%/3.1\%, VideoMMMU by 1.1\%, and TVGBench mIoU by 4.5\%.
It also outperforms VideoRFT and VideoR1 on VideoMME and exceeds them on TVGBench by 6.5\% and 11.2\%, respectively, showing that explicit temporal evidence offers an advantage over text-only reasoning for localization.
Overall, Open-o3-Video-7B remains comparable to other video understanding and reasoning models while additionally providing intuitive spatio-temporal evidence for its predictions.
However, the fixed-context v1 model remains limited in harder temporal regimes. Its performance decreases on LVBench, whose videos average over one hour~\cite{LVBench}, and on Charades-TimeLens~\cite{zhang2026timelens}. This limitation in extreme long-video understanding and fine-grained temporal localization motivates the adaptive clue-seeking mechanism in v2.

\noindent
\textit{Open-o3-Video-v2.}
Open-o3-Video-v2-4B consistently improves over Qwen3-VL-4B across all reported benchmarks.
Its gains of 1.3\%, 0.8\%, and 4.6\% on VideoMME, VideoMME-v2, and WorldSense overall demonstrate stronger general video comprehension, while the 6.3\% improvement in WorldSense recognition indicates enhanced fine-grained visual recognition.
The 5.4\% gain on VideoMMMU further reflects stronger knowledge-intensive video reasoning. On the more challenging Video-Holmes and LVBench, it improves multi-hop clue integration and long-video understanding by 10.7\% and 7.2\%, respectively.
The benefits also extend to temporal grounding: Open-o3-Video-v2-4B improves Charades-TimeLens mIoU by 4.2\% and reaches 31.3\% mIoU on TVGBench, exceeding Open-o3-Video-7B by 10.5\%.
Open-o3-Video-v2-8B achieves the strongest results among our variants on VideoMME, VideoMME-v2, VideoMMMU, LVBench, and Charades-TimeLens.
Across both model scales, the broad gains show that combining evidence-grounded reasoning with agentic clue seeking strengthens fine-grained spatio-temporal localization while improving, rather than sacrificing, general video understanding and reasoning.

\begin{figure*}[t]
    \centering
    \includegraphics[width=\textwidth]{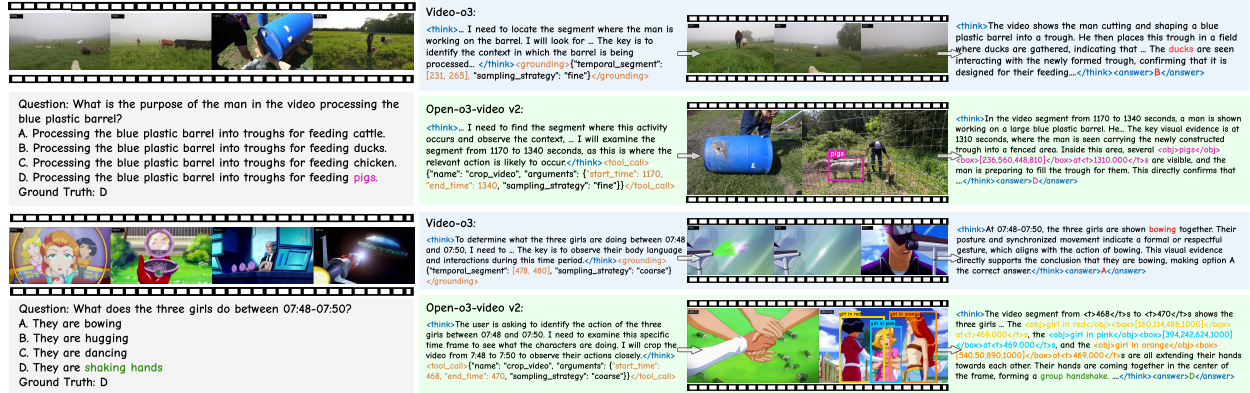}
    \caption{
    Qualitative comparison with Video-o3.
    \textbf{Left:} Video-o3 searches an incorrect segment, while Open-o3-Video-v2 inspects the relevant 1170--1340\,s interval and grounds the pigs at 1310\,s, providing direct evidence for answer D.
    \textbf{Right:} although the question explicitly specifies 07:48--07:50, Video-o3 queries 478--480\,s and incorrectly predicts bowing. Open-o3-Video-v2 inspects the correct 468--470\,s interval, grounds the three girls at 469\,s, and identifies their group handshake.
    In both cases, the acquired observation is converted into explicit spatio-temporal evidence and used directly in the subsequent reasoning process.
    }
    \label{fig:main_good_cases}
\end{figure*}

\subsection{Ablation Studies and Analysis}
\label{sec:ablation_analysis}

\begin{table}[t]
  \centering
  \caption{Stage-wise ablation of Open-o3-Video-v2-4B (\%). V-STaR is evaluated with mAM and mLGM, while VideoMME is evaluated with accuracy.}
  \label{tab:ablation_training_stages}
  \setlength{\tabcolsep}{3.5pt}
  \scriptsize
  \begin{tabular}{lccc}
    \toprule[0.15em]
    \textbf{Model} & \multicolumn{2}{c}{\textbf{V-STaR}} & \textbf{VideoMME} \\
    \cmidrule(lr){2-3} \cmidrule(lr){4-4}
     & \textbf{mAM} & \textbf{mLGM} & \textbf{Acc} \\
    \midrule
    Qwen3-VL-4B (baseline) & 31.9 & 43.7 & 65.3 \\
    Open-o3-Video-v2-SFT-4B (SFT) & 38.2 & 53.4 & 63.7 \\
    Open-o3-Video-v2-4B (SFT+RL) & \textbf{40.4} & \textbf{56.3} & \textbf{66.6} \\
    \bottomrule[0.15em]
  \end{tabular}
\end{table}

We report the complete Open-o3-Video data, reward, training-stage, evidence-faithfulness, and test-time scaling ablations in Appendix~\ref{app-sec:v1_ablation}.
We focus the remaining analysis on our v2 components.

\begin{table}[t]
  \centering
  \caption{Comparison of reasoning mechanisms. All variants use the same backbone, source training data, and evaluation setting. Only the reasoning mechanism is changed.}
  \label{tab:ablation_agentic_data_v2}
  \setlength{\tabcolsep}{3pt}
  \scriptsize
  \begin{tabular}{lccc}
    \toprule
    \textbf{Reasoning mechanism} & \textbf{Video-Holmes} & \textbf{LVBench} & \textbf{V-STaR mLGM} \\
    \midrule
    Qwen3-VL-4B (base)~\cite{bai2025qwen3vl} & 37.3 & 46.0 & 43.7 \\
    Single-turn, no evidence & 45.0 & 48.8 & 47.8 \\
    Multi-turn, no evidence & 47.1 & 51.9 & 52.3 \\
    Multi-turn with grounded evidence (ours) & \textbf{48.0} & \textbf{53.2} & \textbf{56.3} \\
    \bottomrule
  \end{tabular}
\end{table}


\noindent
\textbf{Contribution of the two training stages.}
We conduct a stage-wise comparison to examine how supervised agentic trajectories and subsequent RL contribute to Open-o3-Video-v2.
As shown in Table~\ref{tab:ablation_training_stages}, SFT improves the Qwen3-VL-4B base model by 6.3\% mAM and 9.7\% mLGM, showing that multi-turn grounded trajectories effectively establish agentic evidence-seeking and spatio-temporal grounding behaviors.
Meanwhile, accuracy on VideoMME decreases from 65.3\% to 63.7\%, suggesting a specialization trade-off: imitation of densely grounded multi-turn trajectories strengthens structured reasoning behavior but may temporarily affect general video QA capability.
CS-GRPO subsequently improves the SFT model by 2.2\% mAM, 2.9\% mLGM, and 2.9\% VideoMME accuracy. Unlike SFT, which mainly imitates annotated trajectories, RL evaluates sampled trajectories with answer, grounded-thinking, and tool-use rewards, allowing the model to explore more effective reasoning strategies beyond the supervised patterns. This preserves the learned seeking and grounding behaviors while recovering general video reasoning capability.
The final model therefore exceeds the base by 8.5\% mAM, 12.6\% mLGM, and 1.3\% VideoMME accuracy.

\noindent
\textbf{Agentic spatio-temporal grounded reasoning.}
%
To isolate the contribution of the reasoning mechanism, we construct three controlled variants: single-turn reasoning without explicit evidence, multi-turn clue seeking without explicit evidence, and our full multi-turn reasoning with grounded spatio-temporal evidence. All variants use the same Qwen3-VL-4B backbone, the same source training data, and the same evaluation setting.
As shown in Table~\ref{tab:ablation_agentic_data_v2}, introducing multi-turn clue seeking improves Video-Holmes from 45.0\% to 47.1\%, LVBench from 48.8\% to 51.9\%, and V-STaR mLGM from 47.8\% to 52.3\%, demonstrating the benefit of actively acquiring observations beyond the initial visual context.
Under the same multi-turn setting, adding explicit grounded evidence further raises the three results to 48.0\%, 53.2\%, and 56.3\%, respectively. The 4.0\% gain in V-STaR mLGM is particularly notable, showing that explicit grounding provides complementary value beyond temporal clue seeking alone.
Together, these comparisons isolate two complementary effects: multi-turn seeking expands the available observations, while explicit grounding associates the acquired observations with localized evidence used in subsequent reasoning.

\begin{table}[t]
  \centering
  \caption{Comparison of GRPO and CS-GRPO variants across three video understanding benchmarks (\%).}
  \label{tab:ablation_cs_grpo}
  \setlength{\tabcolsep}{3.5pt}
  \scriptsize
  \begin{tabular}{lccc}
    \toprule
    \textbf{Method} & \textbf{Video-Holmes} & \textbf{LVBench} & \textbf{VideoMME} \\
    \midrule
    GRPO & 46.7 & 52.0 & 65.1 \\
    CS-GRPO w/o gain bonus & 47.0 & 52.2 & 66.2 \\
    CS-GRPO w/o penalties & 46.9 & 50.4 & 65.5 \\
    CS-GRPO (ours) & \textbf{48.0} & \textbf{53.2} & \textbf{66.6} \\
    \bottomrule
  \end{tabular}
\end{table}


\noindent
\textbf{CS-GRPO reward components.}
To evaluate the effectiveness of CS-GRPO, we compare GRPO and CS-GRPO variants on Video-Holmes, LVBench, and VideoMME. GRPO uses the same SFT initialization, RL data, multi-turn rollouts, and optimization setup as CS-GRPO, but does not explicitly model how successive tool interactions change the accumulated evidence state. The other variants remove either the gain bonus for newly acquired clues or the penalties for unhelpful retrieval.
As shown in Table~\ref{tab:ablation_cs_grpo}, full CS-GRPO achieves the best performance on all three benchmarks, outperforming GRPO by 1.3, 1.2, and 1.5 points on Video-Holmes, LVBench, and VideoMME, respectively.
Removing either the gain bonus or the retrieval penalties leads to consistent performance degradation, showing that rewarding informative evidence gains and regularizing inefficient tool use are both beneficial.
These results demonstrate the effectiveness of CS-GRPO for optimizing multi-turn evidence acquisition. The reward-weight sensitivity analysis in Appendix~\ref{app-sec:more_analysis} further shows that the improvement does not rely on a narrowly tuned reward configuration.

%

Figure~\ref{fig:main_good_cases} provides qualitative examples of Open-o3-Video-v2 correctly locating relevant temporal evidence and grounding the supporting objects.
More quantitative analyses are provided in Appendix~\ref{app-sec:more_analysis}, including reward-weight sensitivity, the joint effect of initial visual context and interaction budget, matched visual-budget comparisons, the faithfulness of adaptive evidence seeking, long-horizon evaluation analysis, and answer--grounding consistency.
Appendix~\ref{app-sec:qualitative} further presents qualitative comparisons and failure analysis.

\enlargethispage{2\baselineskip}
\section{Conclusion}
\label{sec:conclusion}
We introduce Open-o3-Video-v2, which formulates video reasoning as an agentic, evidence-grounded process. The model actively identifies missing information, acquires targeted observations, and integrates the decisive moments and grounded objects into its subsequent inference. We further develop STGR-v2-CoT-135K, STGR-v2-RL-13K, and CS-GRPO to make this behavior effective and efficient. Experiments show consistent improvements in grounded reasoning, general video understanding, and temporal localization. These results demonstrate the effectiveness of jointly seeking, grounding, and reasoning with explicit visual evidence.
%
%
We hope our findings can provide the community with insights for building stronger spatio-temporal reasoning MLLMs.

\noindent
\textbf{Contributors:} Jiahao Meng*, Zhongyuan Bao*, Yue Tan, Haiquan Wen, Renrui Zhang, Anna Wang, Jason Li, Yanwei Fu, Yunhai Tong, and Ming-Hsuan Yang. (First two authors share equal contributions.)

\noindent
\textbf{Acknowledgement.} 
This work is supported by the National Key Research and Development Program of China (No. 2023YFC3807600).

\clearpage
\bibliographystyle{plainnat}
\setlength{\bibhang}{0pt}
\setlength{\bibindent}{0pt}
\bibliography{main}

@String(ECCV= {Eur. Conf. Comput. Vis.})

@String(NeurIPS= {Adv. Neural Inform. Process. Syst.})

@String(ICLR = {Int. Conf. Learn. Represent.})

@String(ECCV  = {ECCV})

@String(NeurIPS  = {NeurIPS})

@String(ICLR  = {ICLR})

@misc{openai-o3,
    title = {{OpenAI-o3}},
    author = {OpenAI},
    year = {2025},
    howpublished = "\url{ https://openai.com/index/introducing-o3-and-o4-mini/}"
}

@inproceedings{zhang2025videorefer,
    title = {{VideoRefer} Suite: Advancing Spatial-Temporal Object Understanding with Video {LLM}},
    author = {Yuqian Yuan and Hang Zhang and Wentong Li and Zesen Cheng and Boqiang Zhang and Long Li and Xin Li and Deli Zhao and Wenqiao Zhang and Yueting Zhuang and Jianke Zhu and Lidong Bing},
    booktitle = {Proceedings of the IEEE/CVF Conference on Computer Vision and Pattern Recognition},
  publisher = {IEEE Computer Society},
  address   = {Los Alamitos, CA, USA},
    pages = {18970-18980},
    year = {2025}
}

@misc{zeng2026video,
    title={Video-o3: Native Interleaved Clue Seeking for Long Video Multi-Hop Reasoning},
    author = {Zeng, Xiangyu and Zhang, Zhiqiu and Zhu, Yuhan and Li, Xinhao and Wang, Zikang and Ma, Changlian and Zhang, Qingyu and Huang, Zizheng and Ouyang, Kun and Jiang, Tianxiang and others},
    howpublished = {arXiv preprint arXiv:2601.23224},
    year = {2026}
}

@inproceedings{videochat2,
  title={MVBench: A Comprehensive Multi-modal Video Understanding Benchmark},
  author={Li, Kunchang and Wang, Yali and He, Yinan and Li, Yizhuo and Wang, Yi and Liu, Yi and Wang, Zun and Xu, Jilan and Chen, Guo and Luo, Ping and others},
  booktitle={Proceedings of the IEEE/CVF Conference on Computer Vision and Pattern Recognition},
  pages={22195--22206},
  year={2024}
}

@article{internvl2.5,
  title={Expanding Performance Boundaries of Open-Source Multimodal Models with Model, Data, and Test-Time Scaling},
  author={Chen, Zhe and Wang, Weiyun and Cao, Yue and Liu, Yangzhou and Gao, Zhangwei and Cui, Erfei and Zhu, Jinguo and Ye, Shenglong and Tian, Hao and Liu, Zhaoyang and others},
  journal={arXiv preprint arXiv:2412.05271},
  year={2024}
}

@misc{liu2026videotemp,
  title={VideoTemp-o3: Harmonizing Temporal Grounding and Video Understanding in Agentic Thinking-with-Videos},
  author={Liu, Wenqi and Wang, Yunxiao and Ma, Shijie and Liu, Meng and Su, Qile and Zhang, Tianke and Fan, Haonan and Liu, Changyi and Jiang, Kaiyu and Chen, Jiankang and others},
  howpublished={arXiv preprint arXiv:2602.07801},
  year={2026}
}

@misc{maaz2025videor2,
  title={Video-R2: Reinforcing Consistent and Grounded Reasoning in Multimodal Language Models},
  author={Maaz, Muhammad and Rasheed, Hanoona and Khan, Fahad Shahbaz and Khan, Salman},
  howpublished={arXiv preprint arXiv:2511.23478},
  year={2025}
}

@misc{zeng2026videoo3,
  title={Video-o3: Native Interleaved Clue Seeking for Long Video Multi-Hop Reasoning},
  author={Zeng, Xiangyu and Zhang, Zhiqiu and Zhu, Yuhan and Li, Xinhao and Wang, Zikang and Ma, Changlian and Zhang, Qingyu and Huang, Zizheng and Ouyang, Kun and Jiang, Tianxiang and others},
  howpublished={arXiv preprint arXiv:2601.23224},
  year={2026}
}

@misc{yan2025videochat,
  title={Videochat-r1. 5: Visual test-time scaling to reinforce multimodal reasoning by iterative perception},
  author={Yan, Ziang and Li, Xinhao and He, Yinan and Yue, Zhengrong and Zeng, Xiangyu and Wang, Yali and Qiao, Yu and Wang, Limin and Wang, Yi},
  howpublished={arXiv preprint arXiv:2509.21100},
  year={2025}
}

@misc{zhang2025thinking,
  title={Thinking with videos: Multimodal tool-augmented reinforcement learning for long video reasoning},
  author={Zhang, Haoji and Gu, Xin and Li, Jiawen and Ma, Chixiang and Bai, Sule and Zhang, Chubin and Zhang, Bowen and Zhou, Zhichao and He, Dongliang and Tang, Yansong},
  howpublished={arXiv preprint arXiv:2508.04416},
  year={2025}
}

@misc{zhang2025rewatch,
  title={ReWatch-R1: Boosting Complex Video Reasoning in Large Vision-Language Models through Agentic Data Synthesis},
  author={Zhang, Congzhi and Wang, Zhibin and Ma, Yinchao and Peng, Jiawei and Wang, Yihan and Zhou, Qiang and Song, Jun and Zheng, Bo},
  howpublished={arXiv preprint arXiv:2509.23652},
  year={2025}
}

@misc{wang2025pixel,
  title={Pixel reasoner: Incentivizing pixel-space reasoning with curiosity-driven reinforcement learning},
  author={Wang, Haozhe and Su, Alex and Ren, Weiming and Lin, Fangzhen and Chen, Wenhu},
  howpublished={arXiv preprint arXiv:2505.15966},
  year={2025}
}

@misc{ouyang2025conan,
  title={Conan: Progressive Learning to Reason Like a Detective over Multi-Scale Visual Evidence},
  author={Ouyang, Kun and Liu, Yuanxin and Yao, Linli and Cai, Yishuo and Zhou, Hao and Zhou, Jie and Meng, Fandong and Sun, Xu},
  howpublished={arXiv preprint arXiv:2510.20470},
  year={2025}
}

@misc{meng2025open,
  title={Open-o3 Video: Grounded Video Reasoning with Explicit Spatio-Temporal Evidence},
  author={Meng, Jiahao and Li, Xiangtai and Wang, Haochen and Tan, Yue and Zhang, Tao and Kong, Lingdong and Tong, Yunhai and Wang, Anran and Teng, Zhiyang and Wang, Yujing and others},
  howpublished={arXiv preprint arXiv:2510.20579},
  year={2025}
}

@misc{wang2025video-thinker,
  title={Video-Thinker: Sparking" Thinking with Videos" via Reinforcement Learning},
  author={Wang, Shijian and Jin, Jiarui and Wang, Xingjian and Song, Linxin and Fu, Runhao and Wang, Hecheng and Ge, Zongyuan and Lu, Yuan and Cheng, Xuelian},
  howpublished={arXiv preprint arXiv:2510.23473},
  year={2025}
}

@misc{yuan2025videoexplorerthinkvideosagentic,
      title={VideoExplorer: Think With Videos For Agentic Long-Video Understanding}, 
      author={Huaying Yuan and Zheng Liu and Junjie Zhou and Hongjin Qian and Yan Shu and Nicu Sebe and Ji-Rong Wen and Zhicheng Dou},
      year={2025},
      eprint={2506.10821},
      archivePrefix={arXiv},
      primaryClass={cs.CV}
}

@misc{ge2025framemind,
  title={FrameMind: Frame-Interleaved Video Reasoning via Reinforcement Learning},
  author={Ge, Haonan and Wang, Yiwei and Chang, Kai-Wei and Wu, Hang and Cai, Yujun},
  howpublished={arXiv preprint arXiv:2509.24008},
  year={2025}
}

@misc{yang2025longvt,
  title={LongVT: Incentivizing" Thinking with Long Videos" via Native Tool Calling},
  author={Yang, Zuhao and Wang, Sudong and Zhang, Kaichen and Wu, Keming and Leng, Sicong and Zhang, Yifan and Qin, Chengwei and Lu, Shijian and Li, Xingxuan and Bing, Lidong},
  howpublished={arXiv preprint arXiv:2511.20785},
  year={2025}
}

@misc{gu2025thinking,
  title={Thinking With Bounding Boxes: Enhancing Spatio-Temporal Video Grounding via Reinforcement Fine-Tuning},
  author={Gu, Xin and Zhang, Haoji and Fan, Qihang and Niu, Jingxuan and Zhang, Zhipeng and Zhang, Libo and Chen, Guang and Chen, Fan and Wen, Longyin and Zhu, Sijie},
  howpublished={arXiv preprint arXiv:2511.21375},
  year={2025}
}

@misc{ding2025videozoomer,
  title={VideoZoomer: Reinforcement-Learned Temporal Focusing for Long Video Reasoning},
  author={Ding, Yang and Zhang, Yizhen and Lai, Xin and Chu, Ruihang and Yang, Yujiu},
  howpublished={arXiv preprint arXiv:2512.22315},
  year={2025}
}

@inproceedings{zhang2025vtimecot,
  title={Vtimecot: Thinking by drawing for video temporal grounding and reasoning},
  author={Zhang, Jinglei and Guo, Yuanfan and Potamias, Rolandos Alexandros and Deng, Jiankang and Xu, Hang and Ma, Chao},
  booktitle={Proceedings of the IEEE/CVF International Conference on Computer Vision},
  publisher = {IEEE Computer Society},
  address   = {Los Alamitos, CA, USA},
  pages={24203--24213},
  year={2025}
}

@misc{feng2025video-r1,
  title={Video-r1: Reinforcing video reasoning in mllms},
  author={Feng, Kaituo and Gong, Kaixiong and Li, Bohao and Guo, Zonghao and Wang, Yibing and Peng, Tianshuo and Wu, Junfei and Zhang, Xiaoying and Wang, Benyou and Yue, Xiangyu},
  howpublished={arXiv preprint arXiv:2503.21776},
  year={2025}
}

@inproceedings{wang2025video-rts,
  title={Video-rts: Rethinking reinforcement learning and test-time scaling for efficient and enhanced video reasoning},
  author={Wang, Ziyang and Yoon, Jaehong and Yu, Shoubin and Islam, Md Mohaiminul and Bertasius, Gedas and Bansal, Mohit},
  booktitle={Proceedings of the 2025 Conference on Empirical Methods in Natural Language Processing},
  publisher = {Association for Computational Linguistics},
  address   = {Stroudsburg, PA, USA},
  pages={28114--28128},
  year={2025}
}

@misc{ouyang2025spacer,
  title={SpaceR: Reinforcing MLLMs in Video Spatial Reasoning},
  author={Ouyang, Kun and Liu, Yuanxin and Wu, Haoning and Liu, Yi and Zhou, Hao and Zhou, Jie and Meng, Fandong and Sun, Xu},
  howpublished={arXiv preprint arXiv:2504.01805},
  year={2025}
}

@misc{wang2025time,
  title={Time-R1: Post-Training Large Vision Language Model for Temporal Video Grounding},
  author={Wang, Ye and Wang, Ziheng and Xu, Boshen and Du, Yang and Lin, Kejun and Xiao, Zihan and Yue, Zihao and Ju, Jianzhong and Zhang, Liang and Yang, Dingyi and others},
  howpublished={arXiv preprint arXiv:2503.13377},
  year={2025}
}

@misc{li2025veripo,
  title={VerIPO: Cultivating Long Reasoning in Video-LLMs via Verifier-Gudied Iterative Policy Optimization},
  author={Li, Yunxin and Chen, Xinyu and Li, Zitao and Liu, Zhenyu and Wang, Longyue and Luo, Wenhan and Hu, Baotian and Zhang, Min},
  howpublished={arXiv preprint arXiv:2505.19000},
  year={2025}
}

@misc{park2025deepvideo,
  title={DeepVideo-R1: Video Reinforcement Fine-Tuning via Difficulty-aware Regressive GRPO},
  author={Park, Jinyoung and Na, Jeehye and Kim, Jinyoung and Kim, Hyunwoo J},
  howpublished={arXiv preprint arXiv:2506.07464},
  year={2025}
}

@misc{huang2025vistadpo,
  title={VistaDPO: Video Hierarchical Spatial-Temporal Direct Preference Optimization for Large Video Models},
  author={Huang, Haojian and Chen, Haodong and Wu, Shengqiong and Luo, Meng and Fu, Jinlan and Du, Xinya and Zhang, Hanwang and Fei, Hao},
  howpublished={arXiv preprint arXiv:2504.13122},
  year={2025}
}

@misc{wang2025videorft,
  title={VideoRFT: Incentivizing Video Reasoning Capability in MLLMs via Reinforced Fine-Tuning},
  author={Wang, Qi and Yu, Yanrui and Yuan, Ye and Mao, Rui and Zhou, Tianfei},
  howpublished={arXiv preprint arXiv:2505.12434},
  year={2025}
}

@misc{dang2025twgrpo,
  title={Reinforcing video reasoning with focused thinking},
  author={Dang, Jisheng and Wu, Jingze and Wang, Teng and Lin, Xuanhui and Zhu, Nannan and Chen, Hongbo and Zheng, Wei-Shi and Wang, Meng and Chua, Tat-Seng},
  howpublished={arXiv preprint arXiv:2505.24718},
  year={2025}
}

@inproceedings{chen2025datasets,
  title={Datasets and recipes for video temporal grounding via reinforcement learning},
  author={Chen, Ruizhe and Luo, Tianze and Fan, Zhiting and Zou, Heqing and Feng, Zhaopeng and Xie, Guiyang and Zhang, Hansheng and Wang, Zhuochen and Liu, Zuozhu and Huaijian, Zhang},
  booktitle={Proceedings of the 2025 Conference on Empirical Methods in Natural Language Processing: Industry Track},
  publisher = {Association for Computational Linguistics},
  address   = {Stroudsburg, PA, USA},
  pages={983--992},
  year={2025}
}

@misc{jin2025cot,
  title={CoT-Vid: Dynamic Chain-of-Thought Routing with Self Verification for Training-Free Video Reasoning},
  author={Jin, Hongbo and Liu, Ruyang and Zhang, Wenhao and Luo, Guibo and Li, Ge},
  howpublished={arXiv preprint arXiv:2505.11830},
  year={2025}
}

@misc{zhi2025videoagent2,
  title={VideoAgent2: Enhancing the LLM-Based Agent System for Long-Form Video Understanding by Uncertainty-Aware CoT},
  author={Zhi, Zhuo and Wu, Qiangqiang and Li, Wenbo and Li, Yinchuan and Shao, Kun and Zhou, Kaiwen and others},
  howpublished={arXiv preprint arXiv:2504.04471},
  year={2025}
}

@misc{yang2024doraemongpt,
  title={Doraemongpt: Toward understanding dynamic scenes with large language models (exemplified as a video agent)},
  author={Yang, Zongxin and Chen, Guikun and Li, Xiaodi and Wang, Wenguan and Yang, Yi},
  howpublished={arXiv preprint arXiv:2401.08392},
  year={2024}
}

@misc{ren2025videorag,
  title={Videorag: Retrieval-augmented generation with extreme long-context videos},
  author={Ren, Xubin and Xu, Lingrui and Xia, Long and Wang, Shuaiqiang and Yin, Dawei and Huang, Chao},
  howpublished={arXiv preprint arXiv:2502.01549},
  year={2025}
}

@inproceedings{wang2025videotree,
  title={Videotree: Adaptive tree-based video representation for llm reasoning on long videos},
  author={Wang, Ziyang and Yu, Shoubin and Stengel-Eskin, Elias and Yoon, Jaehong and Cheng, Feng and Bertasius, Gedas and Bansal, Mohit},
  booktitle={Proceedings of the Computer Vision and Pattern Recognition Conference},
  publisher = {IEEE Computer Society},
  address   = {Los Alamitos, CA, USA},
  pages={3272--3283},
  year={2025}
}

@misc{fei2024video,
  title={Video-of-thought: Step-by-step video reasoning from perception to cognition},
  author={Fei, Hao and Wu, Shengqiong and Ji, Wei and Zhang, Hanwang and Zhang, Meishan and Lee, Mong-Li and Hsu, Wynne},
  howpublished={arXiv preprint arXiv:2501.03230},
  year={2024}
}

@inproceedings{zhang2025video-cot,
  title={Video-cot: A comprehensive dataset for spatiotemporal understanding of videos based on chain-of-thought},
  author={Zhang, Shuyi and Hao, Xiaoshuai and Tang, Yingbo and Zhang, Lingfeng and Wang, Pengwei and Wang, Zhongyuan and Ma, Hongxuan and Zhang, Shanghang},
  booktitle={Proceedings of the 33rd ACM International Conference on Multimedia},
  publisher = {Association for Computing Machinery},
  address   = {New York, NY, USA},
  pages={12745--12752},
  year={2025}
}

@misc{ghazanfari2025chain,
  title={Chain-of-Frames: Advancing Video Understanding in Multimodal LLMs via Frame-Aware Reasoning},
  author={Ghazanfari, Sara and Croce, Francesco and Flammarion, Nicolas and Krishnamurthy, Prashanth and Khorrami, Farshad and Garg, Siddharth},
  howpublished={arXiv preprint arXiv:2506.00318},
  year={2025}
}

@misc{hurst2024gpt4o,
  title={Gpt-4o system card},
  author={Hurst, Aaron and Lerer, Adam and Goucher, Adam P and Perelman, Adam and Ramesh, Aditya and Clark, Aidan and Ostrow, AJ and Welihinda, Akila and Hayes, Alan and Radford, Alec and others},
  howpublished={arXiv preprint arXiv:2410.21276},
  year={2024}
}

@misc{ren2023timechat,
  title   = {TimeChat: A Time-sensitive Multimodal Large Language Model for Long Video Understanding},
  author  = {Ren, Shuhuai and Yao, Linli and Li, Shicheng and Sun, Xu and Hou, Lu},
  howpublished = {arXiv preprint arXiv:2312.02051},
  year    = {2023}
}

@inproceedings{huang2024lita,
  title     = {LITA: Language Instructed Temporal-Localization Assistant},
  author    = {Huang, De-An and Liao, Shijia and Radhakrishnan, Subhashree and Yin, Hongxu and Molchanov, Pavlo and Yu, Zhiding and Kautz, Jan},
  booktitle = {European Conference on Computer Vision (ECCV)},
  publisher = {Springer},
  address   = {Cham, Switzerland},
  numpages = {},
  year      = {2024}
}

@inproceedings{li2025unitime,
  title     = {Universal Video Temporal Grounding with Generative Multi-modal Large Language Models},
  author    = {Li, Zeqian and Di, Shangzhe and Zhai, Zhonghua and Huang, Weilin and Wang, Yanfeng and Xie, Weidi},
  booktitle = {Advances in Neural Information Processing Systems (NeurIPS)},
  publisher = {Curran Associates, Inc.},
  address   = {Red Hook, NY, USA},
  numpages = {},
  year      = {2025},
  note      = {Affiliations: Shanghai Jiao Tong University; ByteDance Seed.}
}

@misc{yuan2025sa2va,
  title={Sa2va: Marrying sam2 with llava for dense grounded understanding of images and videos},
  author={Yuan, Haobo and Li, Xiangtai and Zhang, Tao and Sun, Yueyi and Huang, Zilong and Xu, Shilin and Ji, Shunping and Tong, Yunhai and Qi, Lu and Feng, Jiashi and others},
  howpublished={arXiv preprint arXiv:2501.04001},
  year={2025}
}

@misc{wang2024grounded,
  title={Grounded-videollm: Sharpening fine-grained temporal grounding in video large language models},
  author={Wang, Haibo and Xu, Zhiyang and Cheng, Yu and Diao, Shizhe and Zhou, Yufan and Cao, Yixin and Wang, Qifan and Ge, Weifeng and Huang, Lifu},
  howpublished={arXiv preprint arXiv:2410.03290},
  year={2024}
}

@inproceedings{guo2025trace,
  title     = {TRACE: Temporal Grounding Video LLM via Causal Event Modeling},
  author    = {Guo, Yongxin and Liu, Jingyu and Li, Mingda and Liu, Qingbin and Chen, Xi and Tang, Xiaoying},
  booktitle = {International Conference on Learning Representations (ICLR)},
  publisher = {OpenReview.net},
  address   = {Online},
  numpages = {},
  year      = {2025},
  note      = {Affiliations: School of Science and Engineering, The Chinese University of Hong Kong, Shenzhen; Tencent PCG; Shenzhen Institute of Artificial Intelligence and Robotics for Society (AIRS); Guangdong Provincial Key Laboratory of Future Networks of Intelligence.}
}

@misc{bai2025qwen3vl,
  title         = {Qwen3-VL Technical Report},
  author        = {Bai, Shuai and Cai, Yuxuan and Chen, Ruizhe and Chen, Keqin and Chen, Xionghui and others},
  year          = {2025},
  month         = nov,
  eprint        = {2511.21631},
  archivePrefix = {arXiv},
  primaryClass  = {cs.CV}
}

@misc{VideoMME,
  title={Video-MME: The First-Ever Comprehensive Evaluation Benchmark of Multi-modal LLMs in Video Analysis},
  author={Fu, Chaoyou and Dai, Yuhan and Luo, Yongdong and Li, Lei and Ren, Shuhuai and others},
  howpublished={arXiv preprint arXiv:2405.21075},
  year={2024}
}

@misc{VSTaR,
  title={V-STaR: Benchmarking Video-LLMs on Video Spatio-Temporal Reasoning},
  author={Cheng, Zixu and Hu, Jian and Liu, Ziquan and Si, Chenyang and Li, Wei and others},
  howpublished={arXiv preprint arXiv:2503.11495},
  year={2025}
}

@misc{CGBench,
  title={CG-Bench: Clue-grounded Question Answering Benchmark for Long Video Understanding},
  author={Chen, Guo and Liu, Yicheng and Huang, Yifei and He, Yuping and Pei, Baoqi and others},
  howpublished={arXiv preprint arXiv:2412.12075},
  year={2024}
}

@misc{VideoMMMU,
  title={Video-MMMU: Evaluating Knowledge Acquisition from Multi-Discipline Professional Videos},
  author={Hu, Kairui and Wu, Penghao and Pu, Fanyi and Xiao, Wang and Zhang, Yuanhan and others},
  howpublished={arXiv preprint arXiv:2501.13826},
  year={2025}
}

@misc{WorldSense,
  title={WorldSense: Evaluating Real-world Omnimodal Understanding for Multimodal LLMs},
  author={Hong, Jack and Yan, Shilin and Cai, Jiayin and Jiang, Xiaolong and Hu, Yao and Xie, Weidi},
  howpublished={arXiv preprint arXiv:2502.04326},
  year={2025}
}

@inproceedings{fan2024videoagent,
  title={Videoagent: A memory-augmented multimodal agent for video understanding},
  author={Fan, Yue and Ma, Xiaojian and Wu, Rujie and Du, Yuntao and Li, Jiaqi and Gao, Zhi and Li, Qing},
  booktitle={European Conference on Computer Vision},
  pages={75--92},
  year={2024}
}

@misc{zhang2024llavavideo,
  title={Llava-video: Video instruction tuning with synthetic data},
  author={Zhang, Yuanhan and Wu, Jinming and Li, Wei and Li, Bo and Ma, Zejun and Liu, Ziwei and Li, Chunyuan},
  howpublished={arXiv preprint arXiv:2410.02713},
  year={2024}
}

@inproceedings{han2025videoespresso,
  title={Videoespresso: A large-scale chain-of-thought dataset for fine-grained video reasoning via core frame selection},
  author={Han, Songhao and Huang, Wei and Shi, Hairong and Zhuo, Le and Su, Xiu and Zhang, Shifeng and Zhou, Xu and Qi, Xiaojuan and Liao, Yue and Liu, Si},
  booktitle={Proceedings of the Computer Vision and Pattern Recognition Conference},
  publisher = {IEEE Computer Society},
  address   = {Los Alamitos, CA, USA},
  pages={26181--26191},
  year={2025}
}

@misc{cheng2025videoholmes,
  title        = {Video-Holmes: Can MLLM Think Like Holmes for Complex Video Reasoning?},
  author       = {Cheng, Jiale and Ge, Yuying and Wang, Teng and Ge, Yixiao and Liao, Jingjing and Shan, Ying},
  howpublished = {arXiv preprint arXiv:2505.21374},
  year         = {2025}
}

@misc{VideoMMEv2,
  title        = {Video-MME v2: Evaluating Multimodal LLMs with Cohesive Question Groups on Fresh Videos},
  author       = {{Video-MME Team}},
  howpublished = {Preprint},
  year         = {2025}
}

@misc{TVGBench,
  title   = {Datasets and Recipes for Video Temporal Grounding via Reinforcement Learning},
  author  = {Chen, Ruizhe and Luo, Tianze and Fan, Zhiting and Zou, Heqing and Feng, Zhaopeng and Xie, Guiyang and Zhang, Hansheng and Wang, Zhuochen and Liu, Zuozhu and Huaijian, Zhang},
  howpublished = {arXiv preprint arXiv:2507.18100},
  year    = {2025}
}

@misc{LVBench,
  title   = {LVBench: An Extreme Long Video Understanding Benchmark},
  author  = {Wang, Weihan and Yang, Zehai and Gu, Bin and Hu, Hongcheng and Xu, Ke and Zhang, Jialv and Xu, Jinghui and Ma, Jieyi and Zhang, Shiyu and Xu, Jinbao and others},
  howpublished = {arXiv preprint arXiv:2406.08035},
  year    = {2024}
}

@misc{qwen3.5,
    title  = {{Qwen3.5}: Towards Native Multimodal Agents},
    author = {{Qwen Team}},
    month  = {February},
    year   = {2026},
    url    = {https://qwen.ai/blog?id=qwen3.5}
}

@inproceedings{xu2026omtg,
  title={Towards One-to-Many Temporal Grounding},
  author={Qi Xu and Tan Yue and Shihao Chen and Jiahao Meng and Anran Wang and Shunping Ji and Hao Fei and Xiangtai Li},
  booktitle={Proceedings of the 43rd International Conference on Machine Learning ({ICML})},
  publisher = {PMLR},
  address   = {Brookline, MA, USA},
  numpages = {},
  year={2026}
}

@article{zhao2026videoseeker,
  title={VideoSeeker: Incentivizing Instance-level Video Understanding via Native Agentic Tool Invocation},
  author={Zhao, Yiming and Zeng, Yu and Huang, Wenxuan and Fang, Zhen and Miao, Qing and Su, Qisheng and Zhao, Jiawei and Cai, Jiayin and Chen, Lin and Chen, Zehui and others},
  journal={arXiv preprint arXiv:2605.16079},
  year={2026}
}

@article{an2026llava,
  title={LLaVA-OneVision-2: Towards Next-Generation Perceptual Intelligence},
  author={An, Xiang and Xie, Yin and Tang, Feilong and Yan, Yunyao and Tan, Huajie and Zhu, Didi and Chen, Changrui and Zhao, Xiuwei and Qin, Bin and Yang, Kaicheng and others},
  journal={arXiv preprint arXiv:2605.25979},
  year={2026}
}

@article{zhang2025videollama3,
  title={Videollama 3: Frontier multimodal foundation models for image and video understanding},
  author={Zhang, Boqiang and Li, Kehan and Cheng, Zesen and Hu, Zhiqiang and Yuan, Yuqian and Chen, Guanzheng and Leng, Sicong and Jiang, Yuming and Zhang, Hang and Li, Xin and others},
  journal={arXiv preprint arXiv:2501.13106},
  year={2025}
}

@inproceedings{zhang2026timelens,
  title={Timelens: Rethinking video temporal grounding with multimodal llms},
  author={Zhang, Jun and Wang, Teng and Ge, Yuying and Ge, Yixiao and Li, Xinhao and Wang, Limin},
  booktitle={Proceedings of the IEEE/CVF Conference on Computer Vision and Pattern Recognition},
  pages={10419--10429},
  year={2026}
}

@inproceedings{jin2022stcat,
  title={Embracing Consistency: A One-Stage Approach for Spatio-Temporal Video Grounding},
  author={Jin, Yang and Li, Yongzhi and Yuan, Zehuan and Mu, Yadong},
  booktitle={Advances in Neural Information Processing Systems},
  volume={35},
  pages={29192--29204},
  year={2022}
}

@inproceedings{bhattacharyya2023lrr,
  title={Look, Remember and Reason: Grounded Reasoning in Videos with Language Models},
  author={Bhattacharyya, Apratim and Panchal, Sunny and Lee, Mingu and Pourreza, Reza and Madan, Pulkit and Memisevic, Roland},
  booktitle={International Conference on Learning Representations},
  year={2024}
}

@inproceedings{li2025llava-st,
  title={{LLaVA-ST}: A Multimodal Large Language Model for Fine-Grained Spatial-Temporal Understanding},
  author={Li, Hongyu and Chen, Jinyu and Wei, Ziyu and Huang, Shaofei and Hui, Tianrui and Gao, Jialin and Wei, Xiaoming and Liu, Si},
  booktitle={Proceedings of the Computer Vision and Pattern Recognition Conference},
  pages={8592--8603},
  year={2025}
}

@inproceedings{wang2025traceable,
  title={Traceable Evidence Enhanced Visual Grounded Reasoning: Evaluation and Methodology},
  author={Wang, Haochen and Li, Xiangtai and Huang, Zilong and Wang, Anran and Wang, Jiacong and Zhang, Tao and Zheng, Jiani and Bai, Sule and Kang, Zijian and Feng, Jiashi and others},
  booktitle={International Conference on Learning Representations},
  year={2026}
}

@inproceedings{shao2024viscot,
  title={Visual CoT: Advancing Multi-Modal Language Models with a Comprehensive Dataset and Benchmark for Chain-of-Thought Reasoning},
  author={Shao, Hao and Qian, Shengju and Xiao, Han and Song, Guanglu and Zong, Zhuofan and Wang, Letian and Liu, Yu and Li, Hongsheng},
  booktitle={Advances in Neural Information Processing Systems},
  volume={37},
  pages={8612--8642},
  year={2024}
}

@article{cai2026omnivideo100k,
  title={OmniVideo-100K: A Dataset for Audio-Visual Reasoning through Structured Scripts and Evidence Chains},
  author={Cai, Xinyue and Fu, Chaoyou and Zhang, Yi-Fan and He, Ran and Shan, Caifeng},
  journal={arXiv preprint arXiv:2606.14702},
  year={2026}
}

@article{yang2026caprlplusplus,
  title={CapRL++: Unified Reinforcement Learning with Verifiable Rewards for Dense Image and Video Captioning},
  author={Yang, Penghui and Xing, Long and Dong, Xiaoyi and Zang, Yuhang and Cao, Yuhang and Wang, Yibin and Zhou, Yujie and Bu, Jiazi and Liang, Jianze and Huang, Qidong and others},
  journal={arXiv preprint arXiv:2606.09393},
  year={2026}
}

@article{carion2025sam3,
  title={Sam 3: Segment anything with concepts},
  author={Carion, Nicolas and Gustafson, Laura and Hu, Yuan-Ting and Debnath, Shoubhik and Hu, Ronghang and Suris, Didac and Ryali, Chaitanya and Alwala, Kalyan Vasudev and Khedr, Haitham and Huang, Andrew and others},
  journal={arXiv preprint arXiv:2511.16719},
  year={2025}
}

@inproceedings{ravi2025sam2,
  title={Sam 2: Segment anything in images and videos},
  author={Ravi, Nikhila and Gabeur, Valentin and Hu, Yuan-Ting and Hu, Ronghang and Ryali, Chaitanya and Ma, Tengyu and Khedr, Haitham and R{\"a}dle, Roman and Rolland, Chloe and Gustafson, Laura and others},
  booktitle={International Conference on Learning Representations},
  volume={2025},
  pages={28085--28128},
  year={2025}
}

@article{meng2026survey,
  title={Watch, Remember, Reason: Human-View Video Understanding with MLLMs},
  author={Meng, Jiahao and Tan, Yue and Xu, Qi and Gao, Kuan and Liu, Weisong and Li, Yanwei and Li, Jason and Kong, Lingdong and Wang, Haochen and Zhou, Qianyu and others},
  journal={arXiv preprint arXiv:2606.07433},
  year={2026}
}

@article{comanici2025gemini,
  title={Gemini 2.5: Pushing the frontier with advanced reasoning, multimodality, long context, and next generation agentic capabilities},
  author={Comanici, Gheorghe and Bieber, Eric and Schaekermann, Mike and Pasupat, Ice and Sachdeva, Noveen and Dhillon, Inderjit and Blistein, Marcel and Ram, Ori and Zhang, Dan and Rosen, Evan and others},
  journal={arXiv preprint arXiv:2507.06261},
  year={2025}
}

@article{qwen2.5vl,
  title={Qwen2.5-VL Technical Report},
  author={Bai, Shuai and Chen, Keqin and Liu, Xuejing and Wang, Jialin and Ge, Wenbin and Song, Sibo and Dang, Kai and Wang, Peng and Wang, Shijie and Tang, Jun and others},
  journal={arXiv preprint arXiv:2502.13923},
  year={2025}
}

\clearpage
\beginappendix
\noindent
This appendix provides more supporting details and analyses. 
%
%
Appendix~\ref{app-sec:v1_details} describes the data sources, spatio-temporal annotation format, quality-control pipeline and details of the grounded thinking reward used to train Open-o3-Video. 
Appendix~\ref{app-sec:v2_data_details} provides additional details on STGR-v2 annotation quality and human verification. 
Appendix~\ref{app-sec:implementation_details} summarizes the training hyperparameters, video observation settings, and inference configuration of Open-o3-Video-v2.
Appendix~\ref{app-sec:v1_ablation} reports Open-o3-Video stage-wise, reward, data, evidence-faithfulness, and test-time-scaling studies.
Appendix~\ref{app-sec:more_analysis} further analyzes CS-GRPO reward sensitivity, frame and interaction budgets, matched visual-budget performance, adaptive evidence-seeking faithfulness, long-horizon evaluation analysis, and answer--grounding consistency.
Finally, Appendix~\ref{app-sec:qualitative} presents analyses of successful and failed cases of multi-turn evidence acquisition and grounding.

\section{Open-o3-Video Data Construction and Reward Details}
\label{app-sec:v1_details}

\subsection{Data Sources and Composition}
Open-o3-Video requires supervision that connects an answer to what visual evidence is used, when it appears, and where the supporting object is located. We therefore construct two corpora: STGR-CoT-30K for supervised fine-tuning (SFT) and STGR-RL-36K for reinforcement learning (RL). The SFT corpus teaches the model to interleave natural-language reasoning with explicit timestamps and object boxes, whereas the RL corpus exposes it to task-specific rewards for answer correctness and grounded thinking. Both corpora combine temporal grounding, spatial grounding, spatio-temporal grounding, and general video question answering. This mixture preserves broad video understanding while introducing explicit evidence supervision, rather than training a specialist that can localize evidence but cannot answer general video questions.

STGR-CoT-30K contains 4.1K TVG-Coldstart temporal-grounding samples~\cite{TVGBench}, 5K TreeVGR-SFT spatial-grounding samples~\cite{wang2025traceable}, 5.9K newly annotated spatio-temporal samples, and 15K Video-R1-CoT general video QA samples~\cite{feng2025video-r1}. The first three components establish temporal, spatial, and joint evidence-grounding behaviors, while the general video QA component maintains diverse reasoning capability. STGR-RL-36K contains 5.2K temporal-grounding samples from Time-R1~\cite{wang2025time} and TVG-RL~\cite{TVGBench}, 5K spatial-grounding samples from VisCoT~\cite{shao2024viscot}, 10.9K spatio-temporal samples, and 15K Video-R1 samples~\cite{feng2025video-r1}. The spatio-temporal RL subset combines our 5.9K annotations with 5K filtered VideoEspresso samples~\cite{han2025videoespresso}, providing complementary supervision from newly annotated and existing grounded-video data.

\subsection{Spatio-Temporal Annotation}
We annotate the new 5.9K samples with Gemini-2.5-Pro~\cite{comanici2025gemini}. Each sample contains a question--answer pair, one to five key frames, one to three object boxes per key frame, and a reasoning trace that connects the evidence to the answer. The key frames identify when decisive evidence appears, and the boxes identify the query-relevant objects within those frames. The reasoning trace is required to cite these annotations when explaining the answer, thereby aligning the textual rationale with machine-verifiable temporal and spatial evidence. Evidence is serialized as
\begin{center}
\small
\begin{tabular}{l}
\texttt{<obj>object</obj><box>[x1,y1,x2,y2]</box>}\\
\texttt{at<t>time</t>s},
\end{tabular}
\end{center}
where the object name identifies the referent, the box gives its spatial extent, and the timestamp gives its temporal position. This representation is readable as part of the reasoning trace and can also be parsed automatically. It therefore supports direct temporal- and spatial-overlap rewards without requiring a separate latent grounding head.

\subsection{Filtering and Consistency Checking}
We apply three levels of filtering to reduce noisy or unverifiable supervision. First, we remove boxes covering more than 80\% of the frame because they provide little localization information. For every remaining box, we crop the region and ask Qwen2.5-VL~\cite{qwen2.5vl} whether it depicts the annotated object; only affirmative matches are retained. Second, we verify structural consistency: every temporal and spatial reference in the reasoning trace must be covered by the corresponding \texttt{key\_frames} and \texttt{key\_items}. Third, we crop each referenced region and use Qwen2.5-VL to check whether the associated reasoning sentence is semantically supported by its visual evidence. Samples with missing references, invalid or oversized boxes, or inconsistent visual--textual evidence are discarded. These checks ensure that the final traces are not merely well formatted, but also grounded in the annotated video content.

\subsection{Grounded Thinking Reward Details}
For completeness, we provide the detailed grounded-thinking reward used in Open-o3-Video. Let $\{\hat{t}_m\}_{m=1}^{M}$ denote the parsed evidence timestamps. For temporal supervision type $\tau_t\in\{\mathrm{Interval},\mathrm{Point},\varnothing\}$, interval supervision gives $S^{\mathrm{gt}}=[s^{\mathrm{gt}},e^{\mathrm{gt}}]$, point supervision gives $\{t^{\mathrm{gt}}_j\}_{j=1}^{J}$, and $\varnothing$ denotes samples without temporal supervision. With $\Delta t_m=\min_j |\hat{t}_m-t^{\mathrm{gt}}_j|$, the temporal reward is
\begin{equation}
r_{\mathrm{t}}(x,y)
=
\begin{cases}
\displaystyle
\frac{1}{M}\sum_{m=1}^{M}
\mathbb{I}\left[s^{\mathrm{gt}}\leq \hat{t}_m\leq e^{\mathrm{gt}}\right],
& \tau_t=\mathrm{Interval},\\[0.7em]
\displaystyle
\frac{1}{M}\sum_{m=1}^{M}
\exp\left(-\frac{\Delta t_m^2}{2\sigma^2}\right),
& \tau_t=\mathrm{Point},\\[0.7em]
0,
& \tau_t=\varnothing.
\end{cases}
\end{equation}
Adaptive Temporal Proximity anneals $\sigma$ from large to small to provide smooth early feedback and stricter late localization.

For a predicted time--object--box claim $(\hat{t}_m,c_m,b_m)$, let $j^\star(m)=\arg\min_j |\hat{t}_m-t^{\mathrm{gt}}_j|$ be the nearest annotated time and $v_m=\mathrm{IoU}(b_m,b^{\mathrm{gt}}_{j^\star(m)})$ the matched box IoU. Temporal Gating computes the spatial reward only when the predicted timestamp is sufficiently close to the annotated time:
\begin{equation}
r_{\mathrm{s}}(x,y)
=\frac{1}{M}\sum_{m=1}^{M}
\mathbb{I}\left[|\hat{t}_m-t^{\mathrm{gt}}_{j^\star(m)}|\leq \delta\right]\cdot v_m,
\end{equation}
where $\delta$ is the temporal gating threshold.

\section{Additional Details on STGR-v2}
\label{app-sec:v2_data_details}

\noindent\textbf{Human audit protocol.}
We randomly sample 500 examples covering different sources and task types, and two annotators independently audit all examples after stages~(2) and~(3).
After stage~(2), they mainly inspect the correctness of object references and bounding-box annotations.
After stage~(3), they further inspect tool-call validity, reasoning quality, and the consistency between reasoning descriptions and the associated spatio-temporal evidence.
After resolving disagreements, 94.0\% and 91.6\% of the audited examples satisfy the corresponding quality criteria after stages~(2) and~(3), respectively, indicating that the annotation and trajectory-construction pipeline is generally reliable.

We categorize the observed errors into five groups:
\textit{(1) object referent mismatches}, where the localized instance does not correspond to the instance referred to by the question;
\textit{(2) inaccurate object boxes}, where the bounding box is excessively large or does not precisely localize the target object;
\textit{(3) invalid tool-call intervals}, where the requested temporal interval is invalid or inconsistent with the valid video range;
\textit{(4) hallucinated reasoning descriptions}, where the reasoning introduces events, actions, or visual details that are not supported by the video clip; and
\textit{(5) reasoning-evidence mismatches}, where the reasoning description is not aligned with the cited spatio-temporal evidence.
Representative error patterns identified during the audits are incorporated into the consistency checking and filtering criteria in stage~(4).
A final manual review confirms that the resulting data quality meets our expectations.

\noindent\textbf{Inter-annotator agreement.}
Inter-annotator agreement is calculated over all 500 audited examples before disagreement resolution.
After stage~(2), the two annotators achieve an agreement of 94.2\% with a Cohen's $\kappa$ of 0.88.
After stage~(3), they achieve an agreement of 92.6\% with a Cohen's $\kappa$ of 0.84.
These results indicate strong inter-annotator consistency across both evidence annotation and trajectory-quality assessment.
Disagreements are resolved through joint re-inspection of the source video, intermediate annotations, and rewritten reasoning trajectory.

\section{Additional Implementation Details}
\label{app-sec:implementation_details}

\begin{table*}[t]
  \centering
  \caption{Training and optimization configurations of Open-o3-Video-v2-4B.}
  \label{app-tab:v2_training_config}
  \setlength{\tabcolsep}{5pt}
  \scriptsize
  \begin{tabular}{lcc}
    \toprule
    \textbf{Setting} & \textbf{SFT} & \textbf{CS-GRPO} \\
    \midrule
    Training framework & ms-swift & verl AgentLoop \\
    Model initialization & Qwen3-VL-4B-Instruct & SFT checkpoint \\
    Training corpus & STGR-v2-CoT-135K & STGR-v2-RL-13K \\
    Training epochs & 1 & 1 \\
    Global batch size & 16 & 32 \\
    Learning rate & $5\times10^{-6}$ & $1\times10^{-6}$ \\
    Minimum learning rate & $1\times10^{-6}$ & -- \\
    Warmup ratio & 0.05 & 0.03 \\
    Rollouts per prompt & -- & 8 \\
    Rollout temperature & -- & 1.0 \\
    Top-$p$ / Top-$k$ & -- & 1.0 / $-1$ \\
    KL regularization & -- & 0.001, actor loss \\
    Reward weights $(w_{\mathrm{acc}},w_{\mathrm{fmt}},w_{\mathrm{thk}},w_{\mathrm{tool}})$
    & -- & $(1.0,\,0.1,\,0.1,\,0.8)$ \\
    Tool reward coefficients $(\lambda_{\mathrm{clue}},\lambda_{\mathrm{gain}},\lambda_{\mathrm{budget}},\lambda_{\mathrm{repeat}},\lambda_{\mathrm{frag}},\lambda_{\mathrm{drop}})$
    & -- & $(1.0,\,0.1,\,0.1,\,0.1,\,0.1,\,0.05)$ \\
    \bottomrule
  \end{tabular}
\end{table*}

\begin{table*}[t]
  \centering
  \caption{Video observation and interaction settings of Open-o3-Video-v2. SFT frame caps are applied to each observation. For CS-GRPO, crop strategies control the maximum number of uniformly sampled frames within the requested interval rather than fixed sampling rates.}
  \label{app-tab:v2_observation_config}
  \setlength{\tabcolsep}{2pt}
  \scriptsize
  \begin{tabular}{lccccc}
    \toprule
    \textbf{Stage} &
    \makecell{\textbf{Initial}\\\textbf{Observation}} &
    \makecell{\textbf{Coarse}\\\textbf{Crop}} &
    \makecell{\textbf{Medium}\\\textbf{Crop}} &
    \makecell{\textbf{Fine}\\\textbf{Crop}} &
    \makecell{\textbf{Interaction}\\\textbf{Limit}} \\
    \midrule
    SFT &
    1 FPS, $\leq96$ frames &
    1 FPS, $\leq96$ frames &
    2 FPS, $\leq96$ frames &
    4 FPS, $\leq96$ frames &
    \makecell{Given by annotated\\trajectory} \\
    CS-GRPO &
    1 FPS, $\leq32$ uniform frames &
    $\leq8$ frames &
    $\leq16$ frames &
    $\leq32$ frames &
    $\leq5$ assistant turns \\
    Inference &
    1 FPS, $\leq96$ frames &
    \multicolumn{3}{c}{Model-selected \texttt{crop\_video} observations} &
    $\leq5$ assistant turns \\
    \bottomrule
  \end{tabular}
\end{table*}

\noindent\textbf{Training configuration.}
Open-o3-Video-v2 follows the two-stage training recipe in Sec.~\ref{sec:v2_training_recipes}.
For the 4B model, SFT starts from Qwen3-VL-4B-Instruct and performs full-parameter optimization on STGR-v2-CoT-135K for one epoch.
We use a global batch size of 16, a maximum sequence length of 64,000, and a learning rate of $5\times10^{-6}$ with a warmup ratio of 0.05.
CS-GRPO is initialized from the SFT checkpoint.
Each training batch contains 32 prompts, and eight trajectories are sampled for each prompt with temperature 1.0.
We use a learning rate of $1\times10^{-6}$ and a warmup ratio of 0.03.
KL divergence is not included in the reward. The actor objective instead uses a low-variance KL loss with coefficient 0.001.
The main optimization settings are summarized in Table~\ref{app-tab:v2_training_config}.

\noindent\textbf{Video observation and interaction.}
The SFT initial observation uniformly covers the complete video at a target rate of 1 FPS with at most 96 sampled frames.
For multi-turn samples, coarse, medium, and fine crops target 1, 2, and 4 FPS, respectively, with at most 96 frames for each observation.
For samples with annotated key frames, these frames can be additionally inserted into the initial observation.
CS-GRPO uses a more compact initial observation with a target rate of 1 FPS and at most 32 uniformly distributed frames.
Its \texttt{crop\_video} implementation does not assign fixed FPS values to coarse, medium, and fine retrieval.
Instead, it uniformly samples at most 8, 16, and 32 frames from the requested interval, respectively.
Each trajectory allows at most five assistant turns.

\noindent\textbf{Inference configuration.}
At inference time, the model starts from a full-video observation sampled at a target rate of 1 FPS with at most 96 frames.
The model can invoke \texttt{crop\_video} to acquire additional observations and is allowed at most five assistant turns.
This interaction budget is an upper bound rather than a fixed number of calls, allowing the model to terminate once sufficient evidence has been accumulated.

\section{Open-o3-Video Ablation Studies}
\label{app-sec:v1_ablation}

\begin{table*}[t]
  \centering
  \begin{minipage}[t]{0.49\textwidth}
    \centering
    \captionof{table}{Stage-wise ablation of Open-o3-Video-7B (\%). V-STaR uses mAM and mLGM; VideoMME uses accuracy.}
    \label{app-tab:v1_training_stages}
    \setlength{\tabcolsep}{3.5pt}
    \scriptsize
    \begin{tabular}{lccc}
      \toprule
      \textbf{Model} & \textbf{mAM} & \textbf{mLGM} & \textbf{VideoMME} \\
      \midrule
      Qwen2.5-VL-7B & 19.3 & 22.4 & 62.4 \\
      Open-o3-Video-SFT-7B & 28.5 & 37.1 & 60.7 \\
      Open-o3-Video-7B (SFT+RL) & \textbf{33.7} & \textbf{46.6} & \textbf{63.6} \\
      \bottomrule
    \end{tabular}
  \end{minipage}\hfill
  \begin{minipage}[t]{0.49\textwidth}
    \centering
    \captionof{table}{Ablation of adaptive temporal proximity (Ada.) and temporal gating (Gat.) in the thinking reward.}
    \label{app-tab:v1_reward}
    \setlength{\tabcolsep}{7pt}
    \scriptsize
    \begin{tabular}{cccc}
      \toprule
      \textbf{Ada.} & \textbf{Gat.} & \textbf{mAM} & \textbf{mLGM} \\
      \midrule
      $\times$ & \checkmark & 33.0 & 45.2 \\
      \checkmark & $\times$ & 32.3 & 44.9 \\
      \checkmark & \checkmark & \textbf{33.7} & \textbf{46.6} \\
      \bottomrule
    \end{tabular}
  \end{minipage}
\end{table*}

\noindent\textbf{Contribution of the two training stages.}
We separate supervised fine-tuning from reward-based optimization to determine how each stage contributes to Open-o3-Video. As shown in Table~\ref{app-tab:v1_training_stages}, SFT improves the Qwen2.5-VL-7B backbone by 9.2\% mAM and 14.7\% mLGM, demonstrating that supervised reasoning traces establish the basic spatio-temporal grounding behavior. RL subsequently improves the SFT model by 5.2\% mAM, 9.5\% mLGM, and 2.9\% VideoMME accuracy. The final model therefore exceeds the backbone by 14.4\% mAM, 24.2\% mLGM, and 1.2\% VideoMME accuracy. Unlike SFT, which primarily teaches the model to imitate annotated traces, RL permits broader response exploration and rewards more precise evidence and reliable answers. The results show that the two stages are complementary: SFT establishes the grounded output pattern, while RL sharpens evidence localization and recovers general video understanding.

\noindent\textbf{Grounded-thinking reward.}
We remove adaptive temporal proximity and temporal gating individually to test whether both mechanisms are necessary for grounded-thinking optimization. Table~\ref{app-tab:v1_reward} shows that, on the V-STaR benchmark, removing adaptive temporal proximity reduces mAM/mLGM by 0.7\%/1.4\%, whereas removing temporal gating reduces them by 1.4\%/1.7\%. The degradation without temporal gating confirms that spatial overlap should receive credit only when the predicted moment is temporally valid; otherwise, a box at an irrelevant time can introduce a misleading spatial reward. The gains from adaptive temporal proximity validate the complementary role of gradually tightening the temporal criterion: a broader early reward neighborhood alleviates reward sparsity, while the stricter later criterion preserves temporal precision. Together, the two mechanisms provide accessible early supervision without weakening the final grounding objective.

\begin{table*}[t]
  \centering
  \begin{minipage}[t]{0.49\textwidth}
    \centering
    \captionof{table}{Effect of spatio-temporal training data on Open-o3-Video, evaluated on V-STaR (\%).}
    \label{app-tab:v1_data}
    \scriptsize
    \begin{tabular}{lcc}
      \toprule
      \textbf{Training data} & \textbf{mAM} & \textbf{mLGM} \\
      \midrule
      w/o spatio-temporal data & 28.3 & 36.2 \\
      + VideoEspresso & 31.1 & 43.6 \\
      + Our annotations & \textbf{33.7} & \textbf{46.6} \\
      \bottomrule
    \end{tabular}
  \end{minipage}\hfill
  \begin{minipage}[t]{0.49\textwidth}
    \centering
    \captionof{table}{Faithfulness test on a 150-question VideoMME subset. Both interventions remove the same number of frames.}
    \label{app-tab:v1_faithfulness}
    \scriptsize
    \begin{tabular}{lc}
      \toprule
      \textbf{Setting} & \textbf{Accuracy (\%)} \\
      \midrule
      Uniform sampling (64 frames) & \textbf{68.7} \\
      Remove predicted-evidence frames & 66.0 $(-2.7)$ \\
      Remove random frames & 68.0 $(-0.7)$ \\
      \bottomrule
    \end{tabular}
  \end{minipage}
\end{table*}

\noindent\textbf{Spatio-temporal supervision.}
This experiment evaluates whether explicit spatio-temporal training data is necessary and whether our annotation pipeline adds value beyond existing grounded-video supervision. Starting from the setting without spatio-temporal data, adding filtered and rewritten VideoEspresso~\cite{han2025videoespresso} samples improves mAM/mLGM by 2.8\%/7.4\%, as shown in Table~\ref{app-tab:v1_data}. Adding our annotations yields a further 2.6\%/3.0\%, producing total gains of 5.4\% mAM and 10.4\% mLGM. The first improvement demonstrates the importance of joint temporal and spatial supervision, while the additional gain verifies the benefit of our evidence-aligned annotation and filtering pipeline. These results establish high-quality spatio-temporal data as a central component of evidence-grounded video reasoning.

\noindent\textbf{Faithfulness of explicit evidence.}
We evaluate whether the generated spatio-temporal evidence is genuinely related to the final prediction through a controlled frame-removal test. Open-o3-Video produces 1.15 boxes and 1.37 timestamps per instance on average. For every example in a 150-question VideoMME subset, we remove the two frames temporally closest to each predicted timestamp and measure the resulting answer accuracy. As a control, we remove the same number of randomly selected frames. Table~\ref{app-tab:v1_faithfulness} shows that removing evidence-related frames decreases accuracy by 2.7\%, whereas random removal decreases it by only 0.7\%. Because the intervention size is matched, the larger degradation from evidence-frame removal indicates that the cited timestamps capture visual information materially related to the model's answer rather than arbitrary moments in the video.

\begin{table}[t]
  \centering
  \caption{Test-time scaling of Open-o3-Video on WorldSense and VideoMMMU (\%). Confidence-aware voting uses grounded evidence to weight eight sampled responses.}
  \label{app-tab:v1_tts}
  \scriptsize
  \begin{tabular}{lcc}
    \toprule
    \textbf{Setting} & \textbf{WorldSense} & \textbf{VideoMMMU} \\
    \midrule
    Single response ($N=1$) & 37.5 & 52.3 \\
    Majority voting ($N=8$) & 37.3 & 53.1 \\
    Confidence-aware voting ($N=8$) & \textbf{38.5} & \textbf{54.1} \\
    \bottomrule
  \end{tabular}
\end{table}

\begin{table}[t]
  \centering
  \caption{Consistency between answer correctness and spatio-temporal localization on V-STaR Chain1 (\%). Spatial correctness uses vIoU$>0.1$, and temporal correctness uses tIoU$>0.3$.}
  \label{app-tab:qa_grounding_consistency}
  \setlength{\tabcolsep}{3.2pt}
  \scriptsize
  \begin{tabular}{lrrrrrrr}
    \toprule
    \textbf{Model} & \textbf{QA} & \textbf{S} & \textbf{T} & \textbf{S\&T} & \textbf{QA\&S} & \textbf{QA\&T} & \textbf{QA\&S\&T} \\
    \midrule
    Qwen3-VL-4B & 59.6 & 52.7 & 25.9 & 17.8 & 40.1 & 15.6 & 11.7 \\
    Open-o3-Video-v2-4B & 62.8 & 68.0 & 41.5 & 32.8 & 50.1 & 26.4 & 21.7 \\
    \bottomrule
  \end{tabular}
\end{table}

\begin{figure}[t]
  \centering
  \includegraphics[width=0.5\textwidth]{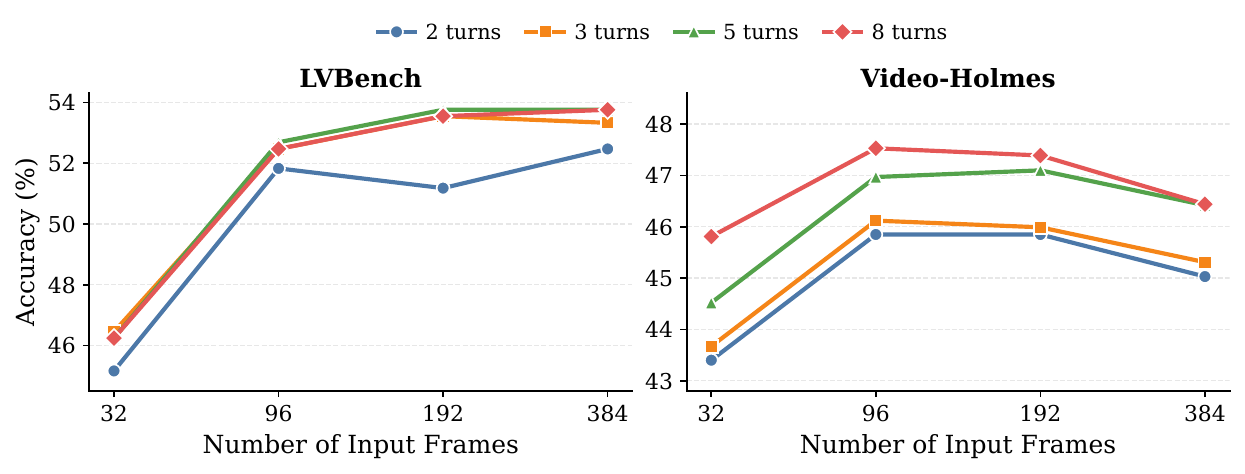}
  \caption{Effect of initial input-frame count and maximum interaction budget on LVBench and Video-Holmes. Results are evaluated on randomly sampled 30\% subsets of each benchmark using the same model and decoding configuration.}
  \label{app-fig:frames_turns_tradeoff}
\end{figure}

\noindent\textbf{Confidence-aware test-time scaling.}
We sample $N=8$ independent responses in parallel with temperature 1.0. Each response contains object names, timestamps, and bounding boxes in its reasoning trace. We extract every predicted box, crop the corresponding region from the original frame, and pair the crop with the question. Open-o3-Video then assigns each crop an evidence score $s\in\{0,1,2\}$, where 0 denotes irrelevant evidence, 1 denotes partially useful evidence, and 2 denotes evidence that strongly supports an answer. We average the scores of all cited objects within a response and use the result as its voting weight. The answer with the highest total weight is selected. This procedure suppresses responses supported by irrelevant or hallucinated regions and favors answers backed by consistent visual evidence.

Table~\ref{app-tab:v1_tts} shows that confidence-aware voting improves the single-response result by 1.0\% on WorldSense and 1.8\% on VideoMMMU. It also exceeds naive majority voting by 1.2\% and 1.0\%, respectively. These results show that explicit spatio-temporal evidence supports inference-time verification and self-correction in addition to making individual predictions interpretable.

\section{More Analysis}
\label{app-sec:more_analysis}

\begin{table*}[t]
  \centering
  \caption{Sensitivity of CS-GRPO to reward weights. Each coefficient is varied independently while all others remain at their default values. Results follow the low/default/high order of the tested weights.}
  \label{app-tab:reward_sensitivity}
  \setlength{\tabcolsep}{5pt}
  \scriptsize
  \begin{tabular}{lllcc}
    \toprule
    \textbf{Reward level} & \textbf{Coefficient} & \textbf{Low / default / high} & \textbf{Video-Holmes} & \textbf{V-STaR mLGM} \\
    \midrule
    \multirow{4}{*}{Final reward $R$}
      & $w_{\mathrm{acc}}$  & $0.5 / 1.0 / 1.5$ & 47.6 / 48.0 / 48.0 & 55.8 / 56.3 / 56.2 \\
      & $w_{\mathrm{fmt}}$  & $0.05 / 0.1 / 0.2$ & 48.1 / 48.0 / 48.2 & 56.1 / 56.3 / 55.7 \\
      & $w_{\mathrm{thk}}$  & $0.05 / 0.1 / 0.2$ & 48.4 / 48.0 / 47.6 & 55.9 / 56.3 / 56.6 \\
      & $w_{\mathrm{tool}}$ & $0.5 / 0.8 / 1.0$ & 47.3 / 48.0 / 47.9 & 56.0 / 56.3 / 56.3 \\
    \midrule
    \multirow{6}{*}{Tool reward $r_{\mathrm{tool}}$}
      & $\lambda_{\mathrm{clue}}$   & $0.5 / 1.0 / 1.5$    & 47.6 / 48.0 / 48.1 & 56.0 / 56.3 / 56.6 \\
      & $\lambda_{\mathrm{gain}}$   & $0.05 / 0.1 / 0.2$   & 47.3 / 48.0 / 47.8 & 55.4 / 56.3 / 57.1 \\
      & $\lambda_{\mathrm{budget}}$ & $0.05 / 0.1 / 0.2$   & 48.2 / 48.0 / 48.6 & 55.7 / 56.3 / 56.0 \\
      & $\lambda_{\mathrm{repeat}}$ & $0.05 / 0.1 / 0.2$   & 47.5 / 48.0 / 47.7 & 56.3 / 56.3 / 56.1 \\
      & $\lambda_{\mathrm{frag}}$   & $0.05 / 0.1 / 0.2$   & 47.6 / 48.0 / 47.5 & 56.7 / 56.3 / 56.6 \\
      & $\lambda_{\mathrm{drop}}$   & $0.025 / 0.05 / 0.1$ & 47.8 / 48.0 / 48.1 & 56.0 / 56.3 / 55.4 \\
    \bottomrule
  \end{tabular}
\end{table*}

\noindent\textbf{Reward sensitivity.}
We study the sensitivity of CS-GRPO at two levels. The first controls the balance among answer correctness, format compliance, grounded thinking, and tool use in the final objective $R$, whose default configuration is $(w_{\mathrm{acc}},w_{\mathrm{fmt}},w_{\mathrm{thk}},w_{\mathrm{tool}})=(1.0,0.1,0.1,0.8)$. The second controls the individual terms within $r_{\mathrm{tool}}$: terminal clue quality, step-wise evidence gain, excess calls, repeated intervals, sequential fragmentation, and evidence degradation. Their default coefficients are $(\lambda_{\mathrm{clue}},\lambda_{\mathrm{gain}},\lambda_{\mathrm{budget}},\lambda_{\mathrm{repeat}},\lambda_{\mathrm{frag}},\lambda_{\mathrm{drop}})=(1.0,0.1,0.1,0.1,0.1,0.05)$.

For each row in Table~\ref{app-tab:reward_sensitivity}, we vary only the indicated coefficient while keeping the remaining reward weights and all training settings unchanged.
Overall, CS-GRPO remains relatively stable across the tested reward ranges, with only modest variations on both Video-Holmes and V-STaR. This suggests that the method does not rely on a narrowly tuned reward configuration. At the same time, several components benefit when their weights increase from very small values to the default setting, indicating that both evidence-gain rewards and penalties on inefficient tool use play meaningful roles in optimization. Overall, the results demonstrate that CS-GRPO is robust to moderate reward-weight changes without requiring precise tuning.

\begin{table}[t]
  \centering
  \caption{Analysis of adaptive evidence acquisition under comparable visual observation budgets (\%). The uniform baseline uses the same model with tool use disabled and uniformly samples 120 frames. For random retrieval, each tool-returned clip is replaced by a length-matched random interval from the same video while preserving the number of tool calls and returned frames.}
  \label{app-tab:adaptive_acquisition_controls}
  \setlength{\tabcolsep}{3.5pt}
  \scriptsize
  \begin{tabular}{lccc}
    \toprule
    \textbf{Setting} & \textbf{VideoMME} & \textbf{Video-Holmes} & \textbf{LVBench} \\
    \midrule
    Uniform sampling (120 frames, no tool) & 65.4 & 47.4 & 45.5 \\
    Length-matched random intervals & 64.6 & 47.1 & 46.2 \\
    Open-o3-Video-v2 & \textbf{66.6} & \textbf{48.0} & \textbf{53.2} \\
    \bottomrule
  \end{tabular}
\end{table}

\noindent\textbf{Input frames and interaction budget.}
We study how the initial observation budget interacts with the maximum number of reasoning turns. We randomly sample 30\% of LVBench and Video-Holmes and evaluate all configurations on the same subsets. We vary the initial observation from 32 to 384 uniformly sampled frames and the maximum interaction budget from 2 to 8 turns, while keeping the model and decoding configuration fixed.

As shown in Fig.~\ref{app-fig:frames_turns_tradeoff}, increasing the initial frame budget from 32 to 96 yields the largest improvement on both benchmarks. With 5 turns, accuracy on LVBench increases from 46.2\% to 52.7\%, while accuracy on Video-Holmes improves from 44.5\% to 47.0\%. Beyond 96 frames, the gains become much smaller: accuracy on LVBench largely saturates around 53--54\%, while accuracy on Video-Holmes changes little between 96 and 192 frames and slightly declines at 384 frames. This suggests that, with active evidence seeking, strong performance does not require an excessively dense initial observation. Even on LVBench, whose videos are typically very long, a moderate initial frame budget can provide sufficient global context for subsequent targeted seeking and grounding.

Increasing the interaction budget is also beneficial, but the marginal gain diminishes as more turns are allowed. LVBench largely saturates by 5 turns, and Video-Holmes shows a similar pattern of progressively smaller improvements at larger budgets. This indicates that the model can often resolve missing evidence within a limited number of interactions, rather than relying on long trajectories with many potentially redundant tool calls. Overall, the results support a moderate initial observation budget together with a bounded interaction horizon, where active seeking complements sparse global observation without requiring either dense frame input or excessive tool use.

At inference time, Open-o3-Video-v2 uses only 2.2 assistant turns on average, including the final answer turn, and processes approximately 120 frames per question.
To isolate the benefit of adaptive evidence seeking, we disable tool use and uniformly sample 120 frames over the full video, matching the average visual observation budget of Open-o3-Video-v2.
As shown in Table~\ref{app-tab:adaptive_acquisition_controls}, the full model improves VideoMME, Video-Holmes, and LVBench by 1.2, 0.6, and 7.7 points, respectively, under a comparable frame budget.
The gap is largest on LVBench, where uniform sampling becomes increasingly sparse as video duration grows, while multi-turn retrieval can selectively zoom into informative temporal intervals with denser observations.
These results indicate that, under comparable visual observation budgets, multi-turn adaptive seeking uses the available frames more effectively by zooming into task-relevant temporal regions rather than uniformly sampling the full video.


\noindent\textbf{Faithfulness of adaptive evidence seeking.}
To examine whether adaptively retrieved observations provide effective evidence for subsequent reasoning and answer prediction, we replace each retrieved clip with a length-matched random interval from the same video for samples involving tool calls, while keeping the number of calls and returned frames unchanged.
All other inference settings remain fixed.
As shown in Table~\ref{app-tab:adaptive_acquisition_controls}, replacing the retrieved observations reduces accuracy by 2.0\% on VideoMME, 0.9\% on Video-Holmes, and 7.0\% on LVBench.
Since the interaction count and visual observation budget remain unchanged, the performance degradation shows that the adaptively retrieved clips provide useful evidence for subsequent reasoning toward the final answer, rather than serving as incidental or superficial tool-use behavior.
The larger drop on LVBench further highlights the importance of retrieving informative evidence for long-video reasoning.

\begin{table}[t]
  \centering
  \caption{Accuracy on a 149-question long-horizon subset of LVBench, covering long videos with sparse evidence and questions involving event chains, ordinal localization, or repeated-instance tracking.}
  \label{app-tab:long_horizon_subset}
  \setlength{\tabcolsep}{7pt}
  \scriptsize
  \begin{tabular}{lc}
    \toprule
    \textbf{Model} & \textbf{Accuracy (\%)} \\
    \midrule
    Qwen3-VL-4B & 37.6 \\
    Video-o3 & 38.3 \\
    Open-o3-Video-v2 & \textbf{48.3} \\
    \bottomrule
  \end{tabular}
\end{table}

\noindent
\textbf{Long-horizon evidence-chain analysis.}
To further evaluate reasoning over sparse evidence in long videos, we construct a long-horizon subset of LVBench using only question patterns and dataset annotations, without using model predictions.
The subset covers before/after event chains, ordinal localization such as first or last occurrences, and repeated-instance tracking.
We further restrict the subset to videos longer than 45 minutes, resulting in 149 questions with a median video length of 79 minutes.
As shown in Table~\ref{app-tab:long_horizon_subset}, Open-o3-Video-v2 achieves 48.3\% accuracy, outperforming Qwen3-VL-4B and Video-o3 by 10.7\% and 10.0\%, respectively.
These results show that the proposed seek--ground--reason formulation is effective when relevant evidence is temporally sparse and must be located or distinguished across distant or repeated events.

\noindent\textbf{Answer--grounding consistency.}
Table~\ref{app-tab:qa_grounding_consistency} examines whether improvements in answer accuracy and spatio-temporal grounding occur jointly rather than independently. Open-o3-Video-v2-4B increases the joint QA--spatial--temporal success rate from 11.7\% to 21.7\%, an absolute gain of 10.0\%. More importantly, among correctly answered questions, the fraction that is also supported by both spatially and temporally correct evidence increases from 19.6\% to 34.5\%. The gains are also consistent across the pairwise metrics: QA--spatial correctness rises from 40.1\% to 50.1\%, while QA--temporal correctness increases from 15.6\% to 26.4\%. These results show that the improvements are not limited to answer accuracy or localization in isolation. Open-o3-Video-v2 increases the frequency with which a correct answer is accompanied by consistent spatio-temporal evidence, directly supporting our goal of reasoning with verifiable visual support.

\section{Qualitative Results}
\label{app-sec:qualitative}

\subsection{Successful Cases}

\begin{figure*}[t]
  \centering
  \includegraphics[width=1.0\textwidth]{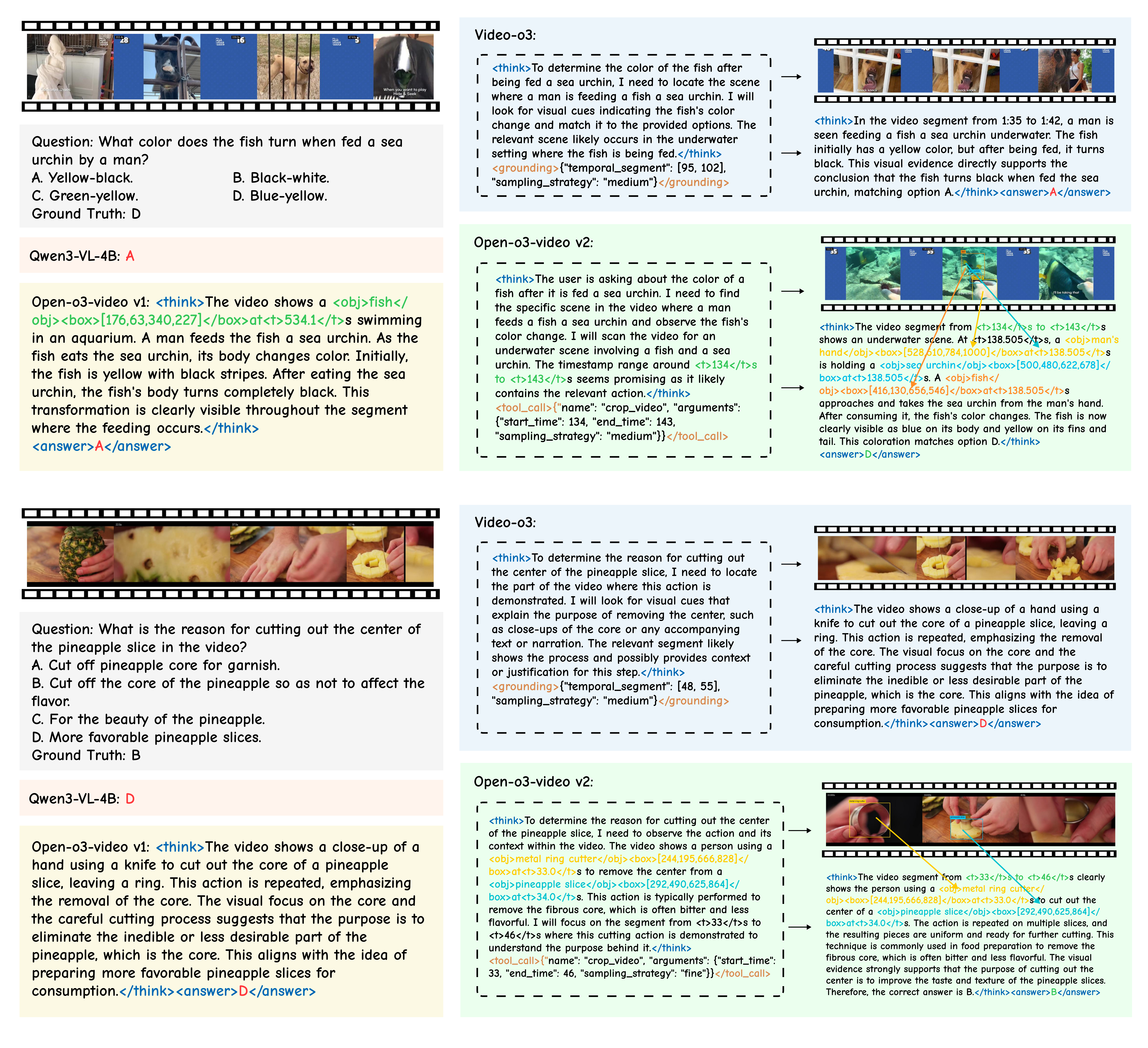}
  \caption{
  Qualitative comparisons on fine-grained visual reasoning.
  \textbf{Top:} For action-change perception, Open-o3-Video-v2 inspects the relevant 134--143\,s interval and grounds the man's hand, sea urchin, and fish at 138.5\,s. The denser observation reveals the fish's blue body and yellow fins after the interaction, enabling the correct answer.
  \textbf{Bottom:} For fine-grained event understanding, Open-o3-Video-v2 grounds the metal ring cutter and pineapple slice and further inspects the 33--46\,s interval using denser temporal sampling. The resulting observation reveals the repeated removal of the fibrous core and provides the missing evidence for answer B.
  In both cases, targeted temporal observation together with explicit object-level grounding exposes the decisive visual evidence and yields a human-verifiable spatio-temporal reasoning trace.
  }
  \label{app-fig:good_case_reasoning}
\end{figure*}

\begin{figure*}[t]
  \centering
  \includegraphics[width=1.0\textwidth]{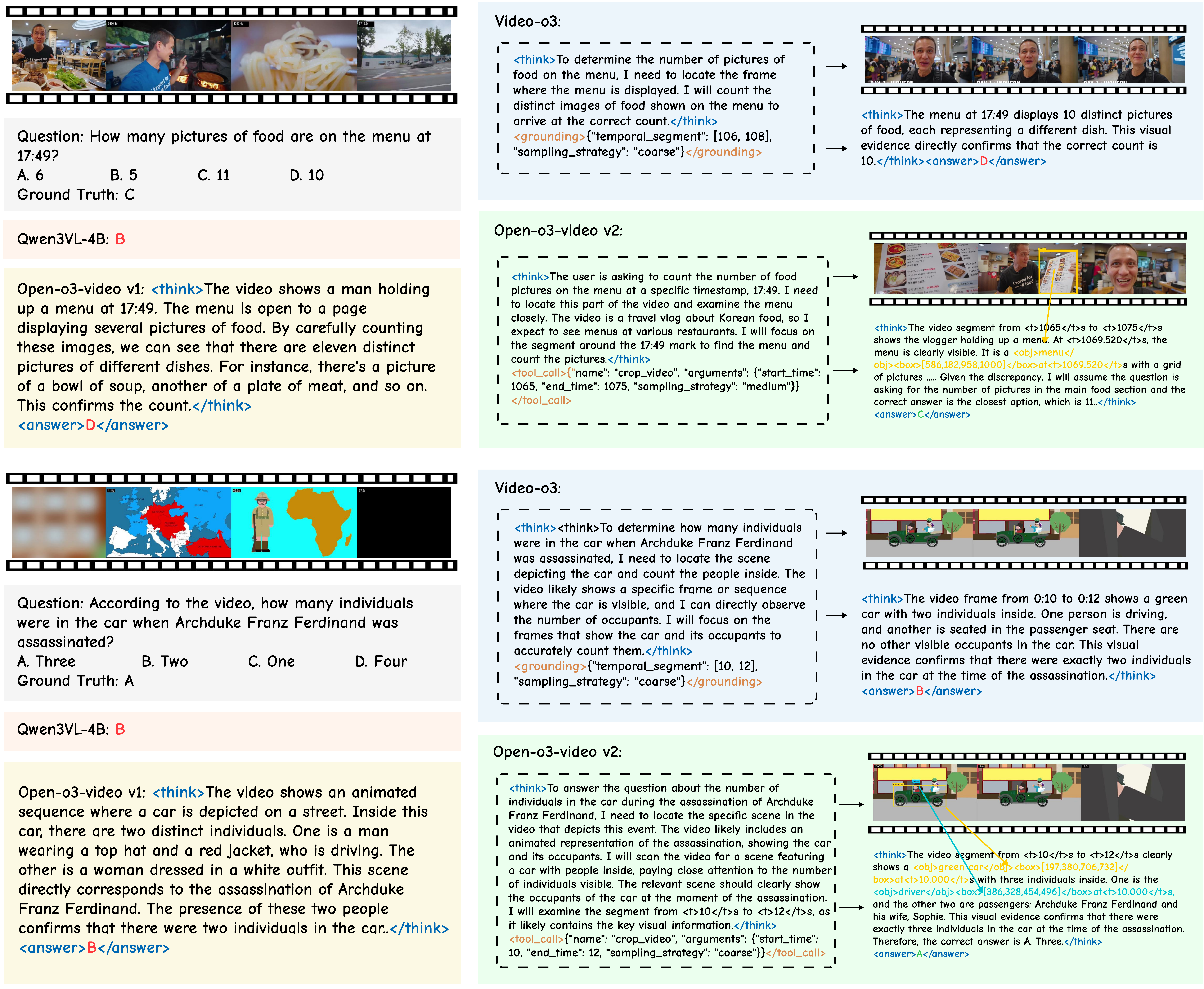}
  \caption{
  Qualitative comparisons on fine-grained visual counting.
  \textbf{Top:} For counting food pictures on the menu at 17:49, Open-o3-Video-v2 inspects the relevant 1065--1075\,s interval and grounds the menu at 1069.5\,s, providing a clearer view of the individual food images and enabling the correct count of 11. In contrast, Video-o3 searches an incorrect temporal interval and produces an incorrect count.
  \textbf{Bottom:} For counting the people in the car, both methods inspect the relevant 10--12\,s interval, but Open-o3-Video-v2 further grounds the car and its occupants, allowing it to distinguish three individuals rather than two.
  In both cases, explicit instance-level grounding complements temporal evidence acquisition by resolving the individual visual entities required for accurate counting.
  }
  \label{app-fig:good_case_counting}
\end{figure*}

Figs.~\ref{app-fig:good_case_reasoning} and~\ref{app-fig:good_case_counting} present representative successful cases of Open-o3-Video-v2.
The first comparison focuses on fine-grained action and event understanding, where the decisive evidence is difficult to resolve from the initial sparse observation.
Open-o3-Video-v2 selectively inspects the relevant temporal intervals and grounds the supporting objects within the acquired observations, exposing visual details that are missed or misinterpreted by the baselines.
The second comparison focuses on fine-grained visual counting, where the target instances are spatially distinguishable once the relevant observation is acquired.
These examples demonstrate how targeted observation and explicit spatio-temporal grounding provide more precise visual evidence for subsequent reasoning.

\begin{figure*}[t]
  \centering
  \includegraphics[width=1.0\textwidth]{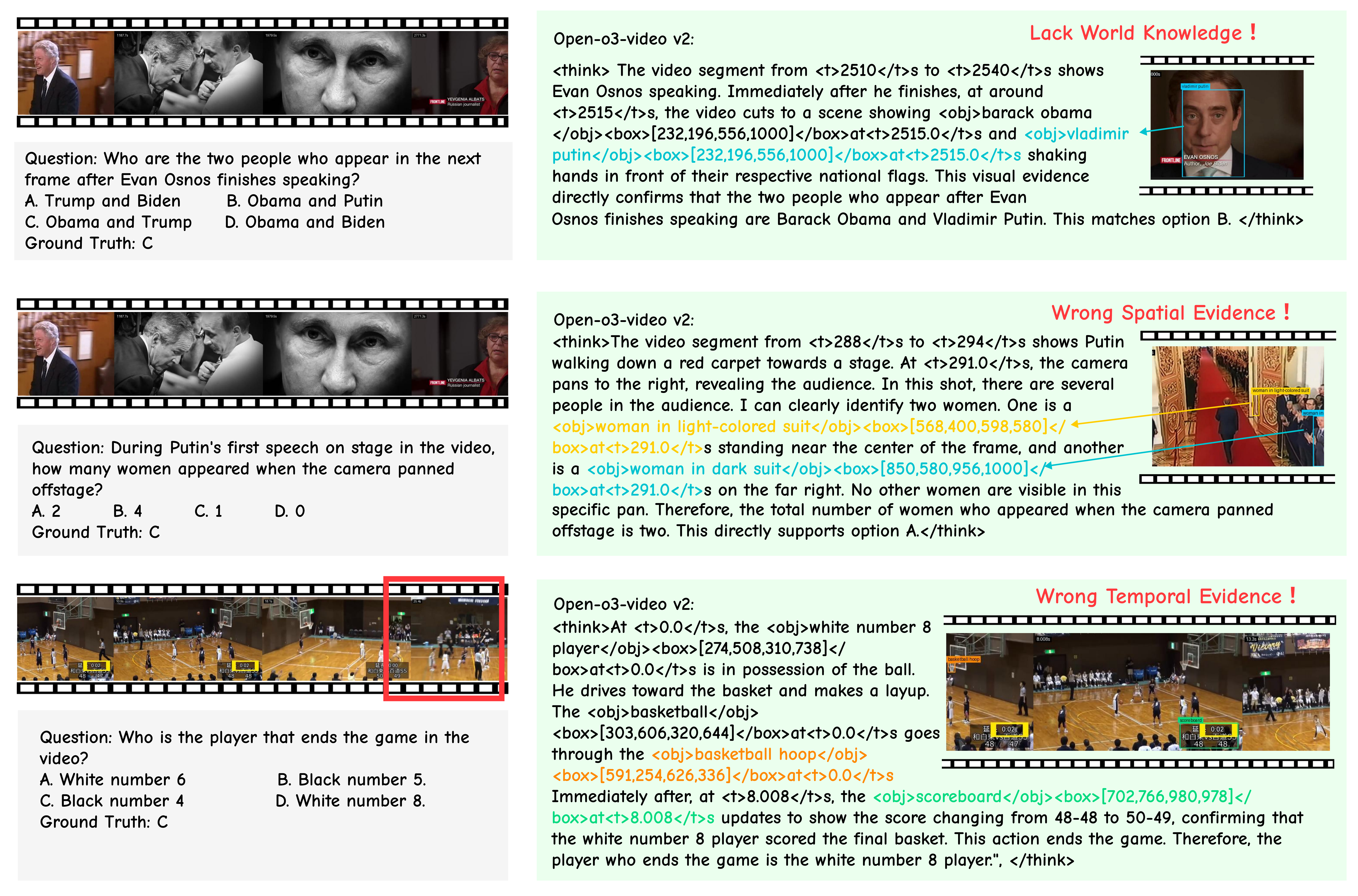}
  \caption{
  Representative failure cases of Open-o3-Video-v2.
  \textbf{Top:} Insufficient identity knowledge leads to an incorrect answer even after a relevant face is retrieved.
  \textbf{Middle:} During a camera pan, inaccurate instance discrimination and association across changing views lead to an incorrect count of the women appearing offstage.
  \textbf{Bottom:} The model anchors its reasoning to the wrong scoring event and misidentifies the player who ends the game.
  These examples expose remaining limitations in world knowledge, fine-grained spatio-temporal instance reasoning, and temporal event selection, respectively.
  }
  \label{app-fig:failure_cases}
\end{figure*}

\subsection{Failure Analysis}

Our quantitative analysis of the incorrect predictions on LVBench shows that precise temporal evidence retrieval and event selection constitute the dominant failure mode, accounting for approximately 43\% of the errors.
Object grounding and fine-grained perception account for another 27\%, including errors in instance discrimination, counting, and visual attribute recognition.
Knowledge-related errors are less frequent, accounting for approximately 6\%, but remain challenging when correct prediction requires identity or external semantic knowledge.
The remaining errors mainly involve reasoning inconsistencies or cases that cannot be attributed to a single factor.

Fig.~\ref{app-fig:failure_cases} illustrates representative examples of these failure modes.
In the first example, the model retrieves a visually relevant face but still fails to identify the public figure correctly, reflecting a remaining limitation in identity and world knowledge.
In the second example, the camera pans across the offstage area, requiring the model to distinguish and associate instances across changing views before determining the total count.
The model instead grounds incorrect instances and produces an erroneous count, indicating that dynamic counting under camera motion remains challenging when visually similar or partially observed objects appear across multiple frames.
In the third example, the model grounds a genuine scoring event but not the game-ending one, indicating difficulty in distinguishing the target event from visually similar events occurring nearby in time.
These failures therefore expose directions for further improvement: better preservation and integration of world knowledge during post-training, and more precise spatio-temporal localization and instance association in complex scenes with confusingly similar events and objects.
Explicit spatio-temporal traces make these weaknesses easier to localize and analyze, while further improvements will require stronger fine-grained perception and temporal event discrimination.

\noindent
\textbf{Future work.} 
%
%
In future work, we will explore longer-horizon video agents that integrate visual evidence with world knowledge and perform counting and analysis on extremely long videos.

\end{document}